\documentclass{article}

\usepackage{microtype}
\usepackage{graphicx}
\usepackage{subfigure}
\usepackage{booktabs} 

\usepackage{hyperref}

\usepackage[accepted]{icml2024}

\usepackage{amsmath}
\usepackage{amssymb}
\usepackage{mathtools}
\usepackage{amsthm}
\usepackage{multirow}
\usepackage{listings}
\usepackage{xcolor}
\usepackage{color}
\usepackage{colortbl}
\usepackage{amsmath}

\usepackage{subcaption}
\usepackage{graphicx}

\usepackage[capitalize,noabbrev]{cleveref}
\usepackage{xspace}

\theoremstyle{plain}

\theoremstyle{definition}

\theoremstyle{remark}

\usepackage[textsize=tiny]{todonotes}

\newcommand{\ours}{{\texttt{PaRe}}\xspace}

\definecolor{mycolor}{rgb}{0.88, 0.13, 0.54}
\colorlet{bestcolor}{mycolor!30!white} 

\definecolor{green}{RGB}{82,208,83}

\icmltitlerunning{Enhancing Cross-Modal Fine-Tuning with Gradually Intermediate Modality Generation}

\begin{document}

\twocolumn[
\icmltitle{Enhancing Cross-Modal Fine-Tuning with \\ Gradually Intermediate Modality Generation}



\icmlsetsymbol{equal}{*}

\begin{icmlauthorlist}
\icmlauthor{Lincan Cai}{bit}
\icmlauthor{Shuang Li}{bit}
\icmlauthor{Wenxuan Ma}{bit}
\icmlauthor{Jingxuan Kang}{uiuc}
\icmlauthor{Binhui Xie}{bit}
\icmlauthor{Zixun Sun}{tent}
\icmlauthor{Chengwei Zhu}{tent}
\end{icmlauthorlist}

\icmlaffiliation{bit}{Beijing Institute of Technology}
\icmlaffiliation{uiuc}{University of Illinois Urbana-Champaign}
\icmlaffiliation{tent}{Interactive Entertainment Group, Tencent}

\icmlcorrespondingauthor{Shuang Li}{shuangli@bit.edu.cn}

\icmlkeywords{Machine Learning, ICML}

\vskip 0.3in
]



\printAffiliationsAndNotice{}  

\begin{abstract}

Large-scale pretrained models have proven immensely valuable in handling data-intensive modalities like text and image. However, fine-tuning these models for certain specialized modalities, such as protein sequence and cosmic ray, poses challenges due to the significant modality discrepancy and scarcity of labeled data. In this paper, we propose an end-to-end method, \ours, to enhance cross-modal fine-tuning, aiming to transfer a large-scale pretrained model to various target modalities. \ours employs a gating mechanism to select key patches from both source and target data. Through a modality-agnostic \textbf{Pa}tch \textbf{Re}placement scheme, these patches are preserved and combined to construct data-rich intermediate modalities ranging from easy to hard. By gradually intermediate modality generation, we can not only effectively bridge the modality gap to enhance stability and transferability of cross-modal fine-tuning, but also address the challenge of limited data in the target modality by leveraging enriched intermediate modality data. Compared with hand-designed, general-purpose, task-specific, and state-of-the-art cross-modal fine-tuning approaches, \ours demonstrates superior performance across three challenging benchmarks, encompassing more than ten modalities.
\end{abstract}

\section{Introduction}

\begin{figure*}[tb]
  \centering
  \includegraphics[width=0.98\textwidth]{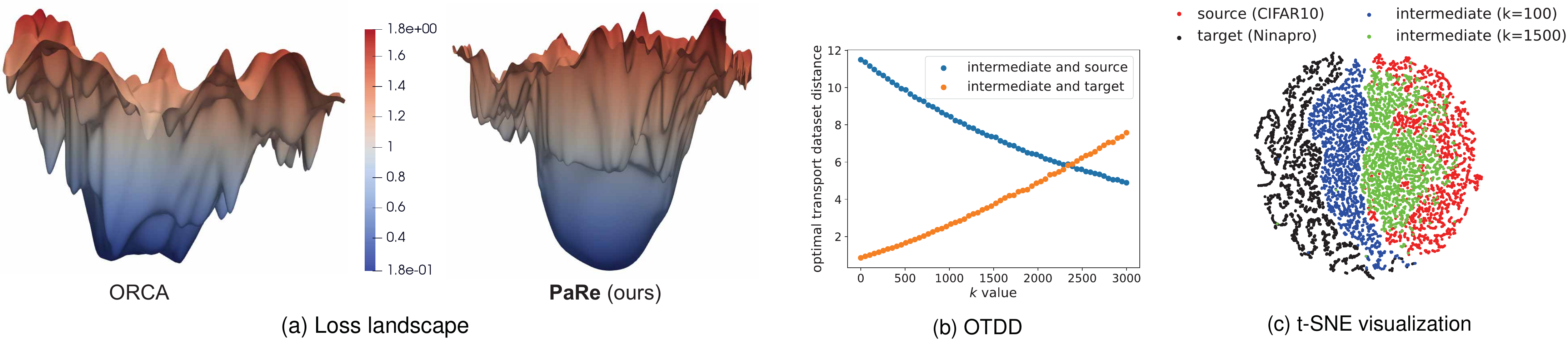}
  \vspace{-0.6em}
  \caption{(a) The loss landscapes of models fine-tuned with ORCA~\cite{orca} and \ours on the Ninapro dataset. (b) The OTDD~\cite{otdd} between the intermediate modality with different $k$ values and source or target modality respectively. (c) Target embeddings (\textcolor{black}{\bf black} dots), intermediate modality embeddings obtained by replacing target patches with different number of source patches (\textcolor{blue}{\bf blue} and \textcolor{green}{\bf green} dots), and source embeddings (\textcolor{red}{\bf red} dots) visualized using t-SNE. Intermediate modalities effectively bridge the modality gap and enhance the model's transferability and stability.}
  \label{Fig_intro}
  \vspace{-1.em}
\end{figure*}
Multimodal perception, as a fundamental component of intelligence, is indispensable to realize artificial general intelligence. Recently, multimodal large language models (MLLMs)~\cite{flamingo,gpt4,team2023gemini,driess2023palm,bai2023qwen,minigpt4,peng2023kosmos,blip2,BaoNXLPWYC0023,dong2023dreamllm} have effectively served as a versatile interface across various tasks encompassing vision, language, and multimodal tasks. While these models demonstrate remarkable performance and broad applicability in conventional modalities such as text, images, and videos, their development demands billions of high-quality data, substantial computational resources, and advanced training techniques. In certain professional field where data is scarce and modality-specific, the need for specialized models becomes apparent.

Fine-tuning contirbutes to this issue by transferring knowledge of large-scale pretrained models to downstream tasks, where data modality usually maintain consistent~\cite{touvron2023llama,RuderH18,PanY10,goyal2023finetune,dehghani2023scaling,jia2022visual,bertasius2021space}. Recent studies have revealed the feasibility of fine-tuning pretrained models for unseen modalities~\cite{moor2023foundation,yang2024touch,orca,lin2023evolutionary,pang2023frozen}. For instance, one could utilize a pretrained vision or language model to tackle genomics tasks. Despite the potential of fine-tuning pretrained models across modalities, cross-modal fine-tuning encounters two challenges: a) \textit{Modality gap}: The gap may arise from task mismatch (e.g., image classification and genetic prediction) and data heterogeneity. Adapting these tasks to facilitate fine-tuning while preserving meaningful representations is arduous. b) \textit{Data scarcity}: Some tasks in certain modalities may suffer from limited labeled data due to the necessity for additional expertise or difficulties in data collection, thus hindering effective fine-tuning.

ORCA~\cite{orca} has undertaken a preliminary exploration to confront the above challenges by employing a two-stage training approach. In the first stage, it enhances model transferability by reducing the optimal transport dataset distance (OTDD)~\cite{otdd} between source and target modality data, aiming to minimize the discrepancy between the two modalities. Subsequently, in the second stage, full fine-tuning is conducted to adapt the pretrained model to the target modality, yielding promising results. Despite its efforts on reducing the OTDD across distinct modalities, the significant modality gap and data scarcity remain to be addressed. As illustrated in Fig.~\ref{Fig_intro}~(a), we show the loss landscape~\cite{landscape} of model fine-tuned with ORCA. It is evident that ORCA faces instability during training, potentially leading to suboptimal results as it is prone to getting trapped in unfavorable local optima. We conjecture that one reason is the alignment stage in ORCA, which can reduce the modality gap but not completely eliminate it. Another factor is that fine-tuning with limited data often leads to overfitting.


Driven by the above analysis, we propose an end-to-end approach to promote cross-modal fine-tuning through gradually generating intermedia modality data and bridging distinct modalities. Motivated by traditional data augmentation techniques like Mixup~\cite{mixup} and CutMix~\cite{cutmix}, we achieve this by mixing source and target data to generate diverse intermediate modalities. However, due to the diversity between source and target modality data, applying mixing operations directly applied in the raw space is not feasible, and the substantial differences between modalities also increase the risk of model confusion.
Here, we thoroughly analyze the characteristics of different modalities and design a modality-agnostic \textbf{Pa}tch \textbf{Re}placement (\ours) method to construct intermediate modalities. Initially, data from source and target modalities are seperately mapped to a unified dimensional space using distinct, specific embedders, facilitating mixing operations at the embedding level. Subsequently, each patch within the source and target samples undergoes scoring through a gate network. A higher score signifies the patch's increased importance and its contribution to the model's classification of the respective sample. Consequently, we select top-$k$ scored source patches to replace bottom-$k$ scored target patches, thereby maximizing the preservation of crucial information in the intermediate modality data for both source and target data.

Take a step further, we facilitate the model's gradual progression from easier intermediate modalities to more challenging ones, ultimately adapting it to the target modality. This curriculum learning process~\cite{bengio2009curriculum} can enhance the stability of the cross-modal fine-tuning. Specifically, we use the OTDD~\cite{otdd} between the generated intermediate modality data and the source modality data as a metric to gauge the difficulty level. As~\citet{otdd} analysed, a smaller OTDD between source and target dataset indicates higher transferability of the source model to the target dataset. The Fig.~\ref{Fig_intro}~(b)-(c) reveal that as the $k$ value increases, the OTDD between the intermediate modality data and the source modality data decreases, enhancing the transferability of the source pretrained model. Thus, we progressively decrease the $k$ value during training to construct intermediate modalities ranging from easy to hard. 
Recall that, as shown in Fig~\ref{Fig_intro}~(a), the loss landscape of the model fine-tuned using \ours appears smoother and is less prone to getting stuck in unfavorable local optima compared to ORCA~\cite{orca}. This phenomenon is attributable to the generation of intermediate modalities and the transition from easy to hard. 

In a nutshell, our contributions are summarized as follows:
\begin{itemize}
\item[$\bullet$] We propose an end-to-end cross-modal fine-tuning framework that is able to adapt a pretrained source model to any target modality. Leveraging the designed gradually intermediate modality generation, one can bridge the modality gap and alleviate the issue of insufficient data in the target modality, enhancing the model's transferability and stability.
\item[$\bullet$] We design a modality-agnostic \textbf{Pa}tch \textbf{Re}placement (\ours) method to construct intermediate modalities. By using a gate network for patch scoring, we extract pivotal patches from embeddings of both source and target modalities, blending them to facilitate a smoother training process with intermediate modalities.
\item[$\bullet$] We validate the effectiveness of \ours on three benchmarks comprising 48 datasets. In the most challenging NAS-360-Bench benchmark which contains 10 modalities, our \ours significantly outperforms other approaches, including task-specific, general-purpose, and cross-modal fine-tuning, across all datasets.
\end{itemize}

\section{Related Work}

\paragraph{Mutil-modal transformers.} Transformers~\cite{Attention} is first used successfully for natural language processing. With the rapid success of large language models (LLMs)~\cite{devlin2018bert,raffel2020exploring,zhang2022opt,BrownMRSKDNSSAA20,touvron2023llama,vicuna2023}, researchers have started aligning multimodal data with LLMs~\cite{clip,flamingo,gpt4,BaoNXLPWYC0023,wang2023image,sun2023generative,hong2023cogagent,chen2023internvl,ye2023mplug,li2023seed,onellm,dong2024internlm,wei2023vary,shukor2023unival}. These general-purpose models excel in perceiving data-rich modalities (e.g., image, video, audio, text), following instructions, and learning in context. However, in many specialized domains where the data is scarced, a well-adapted specialized model is needed.

\paragraph{In-modality fine-tuning.} Pretrained models are widely used in fields like vision (e.g., dense prediction~\cite{kirillov2023segment,liu2021swin} and 3D understanding~\cite{bertasius2021space,luo2022clip4clip}), language (e.g., cross-lingual learning~\cite{Zheng0HWCSC0SW20,YangC0022,lsg} and parameter-efficient~\cite{HuSWALWWC22,HoulsbyGJMLGAG19}), and speech~\cite{radford2023robust,li-etal-2021-multilingual}. This line of methods endeavor to transfer knowledge learned during the pre-training process and data modality of downstream tasks always within seen modalities. But, whether one can transfer from one modality to another irrelevant modality is still under-explored. Imagine harnessing the power of pretrained vision transformers not just for image classification, but for unraveling the intricacies of physics puzzles.

\paragraph{Cross-modality fine-tuning.} Adapting pretrained models to other modalities and tasks has been demonstrated in recent works~\cite{tan2019lxmert,pang2023frozen,gu2022keypoint,orca}. In particular, LLMs have been employed in the life sciences to translate between text and chemistry~\cite{edwards2021text2mol}, biology~\cite{luo2022biogpt}, medical~\cite{moor2023foundation}, DNA-sequencing~\cite{nguyen2023hyenadna}, and protein sequences~\cite{lin2023evolutionary} and folding~\cite{jumper2021highly}). In contrast to the aforementioned approaches, this work aims to fine-tune pretrained models initially trained on general modalities like vision or language, on specialized modalities with scarce data. \citet{orca} have taken the first step in this direction via alignment at the embedding space then fine-tuning the whole network. Different from previous works, our work progressively constructs different intermediate modalities during the training process, which enhances the model's transferability and training stability.

\paragraph{Curriculum learning.} Curriculum Learning~\cite{bengio2009curriculum,zhou2018minimax} promotes the strategy of learning from easier samples first and harder samples later. This idea has been widely explored in training neural networks~\cite{hacohen2019power,wang2023efficienttrain}, reinforcement learning~\cite{NarvekarPLSTS20,klink2022curriculum} and transfer learning~\cite{weinshall2018curriculum,zhang2021flexmatch}. In this work, we intend to narrow the substantial modality gap between source and target data. And we adopt a step-by-step approach, starting with simpler intermediate modalities and gradually moving towards more complex ones. This progressive generation ultimately enhances the stability of the model's cross-modal fine-tuning process to better align with the target modality.

\paragraph{Cross-modality mixing.} Mixup~\cite{mixup} is a commly used and effective technique for data augmentation. There are two main categories: Global Mixup, exemplified by methods like Mixup~\cite{mixup}, Un-Mix~\cite{unmix}, Manifold-Mixup~\cite{manifold} and PatchMix~\cite{patchmix}, and Region Mixup, represented by CutMix~\cite{cutmix}, Saliencymix~\cite{saliencymix} and TransMix~\cite{transmix}. All of these methods only involve Mixup for uni-modal data. As for cross-modality, VLMixer~\cite{vlmixer} employs modality-agnostic augmentation to create semantically invariant cross-modal inputs which proficiently merging visual tokens with non-grounded linguistic tokens. There is semantic correlation among cross-modal data in VLMixer. However, our work focuses on exploring a method akin to Mixup technique for unpaired cross-modal data, where no semantic correlation between modalities and significant modality gap exists.
Through the patch replacement approach we designed, we construct data-rich intermediate modalities, bridging the gap between modalities to enhance the transferability of the source pretrained model. 
\section{Method}

\begin{figure*}[tb]
  \centering
  \includegraphics[width=0.9205\textwidth]{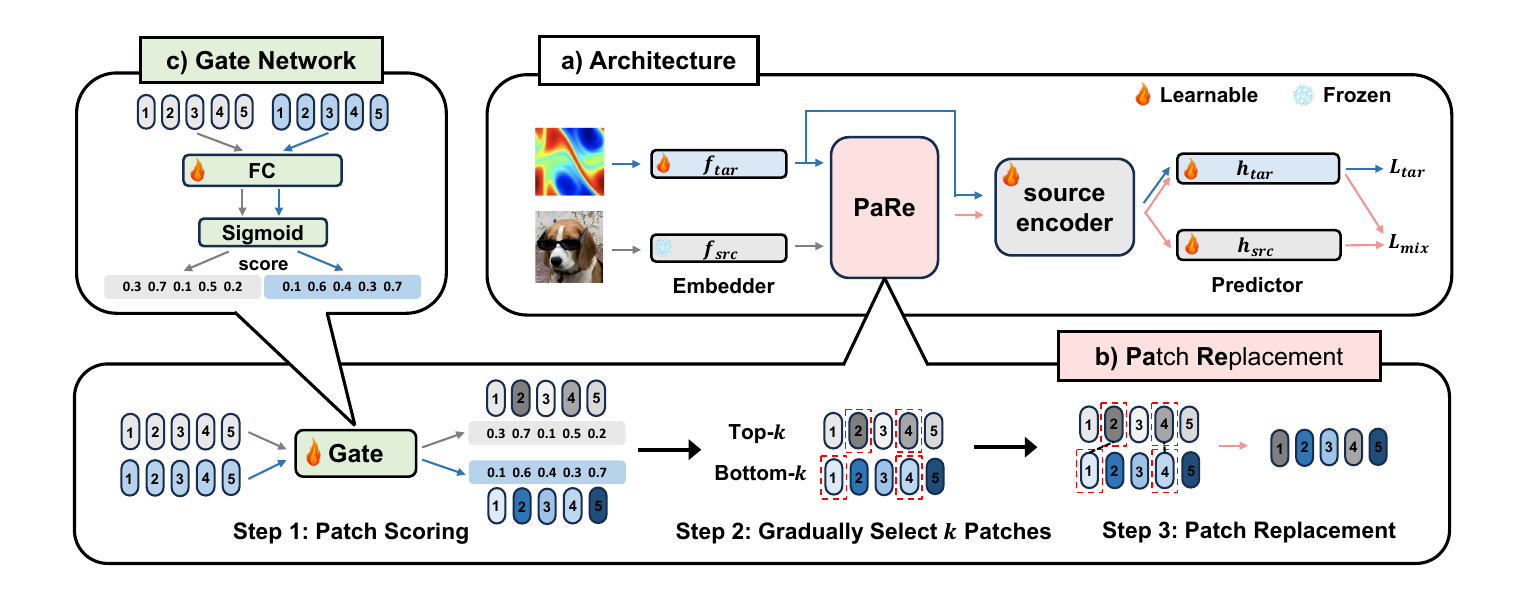}
  \caption{Framework overview. a) The overall architecture of the model and workflow of our method. b) {\bf Pa}tch {\bf Re}placement (\ours) module contains three steps: patch scoring using the designed gate network, gradually select top-$k$ source patches and bottom-$k$ target patches and replace the selected target patches with the source patches one by one. c) The architecture of the gate network which contains a Full-Connected (FC) layer and a Sigmoid layer.}
  \label{Fig_method}
  \vspace{-1.em}
\end{figure*}

\paragraph{Problem setup.}
A modality $\mathcal{M}$ contains a feature space $\mathcal{X}$, a label space $\mathcal{Y}$, and a joint probability distribution $P(\mathcal{X}, \mathcal{Y})$. In this paper, we focus on a more difficult cross-modal setting that the feature space $\mathcal{X}^t$, label space $\mathcal{Y}^t$ and joint probability distribution $P(\mathcal{X}^t, \mathcal{Y}^t)$ in the target modalities $\mathcal{M}^t$ are all different from those in the source (pretrained) modality $\mathcal{M}^s$, i.e., $\mathcal{X}^t\neq\mathcal{X}^s$, $\mathcal{Y}^t\neq\mathcal{Y}^s$, and $P^t(\mathcal{X}^t, \mathcal{Y}^t)\neq P^s(\mathcal{X}^s, \mathcal{Y}^s)$ (e.g., natural images vs. PDEs).

Our goal is to adapt the model pretrained from the source modality $\mathcal{M}^s$ to the target modality $\mathcal{M}^t$ in a supervised manner. In contrast to the two-stage fine-tuning approach employed by ORCA~\cite{orca}, we design an end-to-end fine-tuning approach named \ours. This approach involves constructing intermediate modalities through patch scoring and gradually patch replacement, addressing the gap between the source and target modalities in a progressive way. This approach effectively enhances the model transferabiltiy and training stability, while also alleviating the issue of insufficient training of model due to the limited data for target modality. Fig.~\ref{Fig_method} illustrates the workflow of \ours. In the following subsections, we will provide a detailed overview of each module in \ours.


\subsection{Architecture design}
In order to apply the source modality pretrained models to the target modality, we follow ORCA~\cite{orca}, which decomposes a transformer-based model into three parts: an embedder ${f}$, a feature encoder ${g}$ and a predictor ${h}$, and then employs a pretrained architecture and weights to initialize the feature encoder ${g}$.

\paragraph{Source and target embedder.}
Since our approach requires patch-level replacement of source and target embeddings, serving as inputs to the feature encoder, we need to design source and target-specific embedders to map them to the same dimension. We can denote $f^s$ as the pretrained source embedder, which transforms the source raw data $\mathcal{X}^s$ into sequence source embeddings $\mathcal{\Tilde{X}}^s=\mathbb{R}^{N\times D}$ (where ${N}$ denotes the embeddings length and ${D}$ denotes the embedding dimension). The target embedder $f^t$ is randomly initialized and designed to process target inputs $\mathcal{X}^t$ of arbitrary dimension and transform them to $\mathcal{\Tilde{X}}^t=\mathbb{R}^{N\times D}$. Subsequently, we obtain the mixed embedding $\mathcal{\Tilde{X}}^m$ through Patch Scoring and Patch Replacement, as proposed in Sec~\ref{PaRe}.

\paragraph{Custom predictor.}
Different modalities may correspond to different tasks, various tasks typically entail distinct types of outputs, such as classification logits in $\mathbb{R}^K$, or dense map where the spatial dimension aligns with the input, and per-index logits correspond to $K$ classes. For classification, we follow ORCA~\cite{orca} to utilize average pooling along the sequence length dimension, resulting in 1D tensors with a length of $D$. Subsequently, to make seperate predictions for the feature $g(\mathcal{\Tilde{X}}^t)$ of target modality data and the feature $g(\mathcal{\Tilde{X}}^s)$ of the intermediate modality data, we randomly initialize two classifiers ${h}^t$ and ${h}^s$ to map $D$ to $K$ such as the target prediction $p^{t}=h^t(g(\mathcal{\Tilde{X}}^t))$. In the case of dense prediction tasks, we also follow ORCA's configuration to shape the tensor to the desired output dimension.

\subsection{Gradually intermediate modality generation} \label{img}
Adapting a pretrained source model to various target modalities encounters two issues. The first is the substantial disparity between source and target modalities, which is quantified by the optimal transport dataset distance (OTDD)~\cite{otdd} in this work. A larger OTDD signifies greater divergence between modalities, leading to diminished model transferability. The second issue relates to the inadequate training of the model due to the limited amount of data available in the target modality.

Hence, during the end-to-end training process, we progressively construct intermediate modality data, transitioning from resembling the source modality (easier) to resembling the target modality (harder). Through these intermediate modality data, we bridge the modality gap and alleviate the issue of in-sufficient target data, thereby enhancing the model transferability and the training stability. This method provides a simple yet effective solution to the aforementioned challenges.

\subsection{Modality-agnostic patch replacement} \label{PaRe}
Motivated by in-modality transfer learning, constructing an intermediate domain serves as an effective approach to bridge the domain gap. They often linearly combine images from different domains (such as Mixup~\cite{mixup}) to generate intermediate domain data.

Nevertheless, creating intermediate modality data proves to be more challenging for diverse modalities. The varying input dimensions between target and source modality data hinder the application of methods such as Mixup~\cite{mixup} or CutMix~\cite{cutmix} in raw space. While it may be intuitive to consider these approaches in a unified embedding space, the substantial differences between modalities increase the risk of model confusion through direct Mixup. Furthermore, employing region-based replacement methods like CutMix may not produce favorable outcomes for diverse modality data. The reason is that unlike image modality data where semantic information is often concentrated in the middle region, some modalities, such as PDE data, may harbor more critical information at the edges. Consequently, we opt for a modality-agnostic patch replacement approach to construct intermediate modalities which can maximize the preservation of key information in the intermediate modality for both source and target modalities.

\textbf{Patch scoring with gate network.}
For two modalities without semantic correlation, the most straightforward way of patch replacement is to randomly select patches for replacement. But, random patch replacement may lead to instances where non-semantic patches from source modality data replace crucial patches from target modality data, disrupting the model's training. Therefore, we design a patch selected strategy called \textbf{Patch Scoring} using a \textbf{Gate Network} such that, in the mixed embeddings after the patch replacement, crucial information from both source and target data is well preserved. This preservation is essential for effective model classification and efficient training.

Formally, we denote the source embeddings $\mathcal{\Tilde{X}}^s=\mathbb{R}^{N\times D}$ contains ${N}$ patches that $\mathcal{\Tilde{X}}^s=\{{{\Tilde{x}}^s}_1, {{\Tilde{x}}^s}_2,..., {{\Tilde{x}}^s}_N\}$ and the target embeddings $\mathcal{\Tilde{X}}^t=\mathbb{R}^{N\times D}$ contains ${N}$ patches that $\mathcal{\Tilde{X}}^t=\{{{\Tilde{x}}^t}_1, {{\Tilde{x}}^t}_2,..., {{\Tilde{x}}^t}_N\}$. We score each patch ${{\Tilde{x}}^s}_i$ from source and ${{\Tilde{x}}^t}_i$ from target using a gate network with a fully-connected (FC) layer and a sigmoid ($\sigma$) layer $\mathcal{S}^s=\sigma(FC(\mathcal{\Tilde{X}}^s))$, $\mathcal{S}^t=\sigma(FC(\mathcal{\Tilde{X}}^t))$, the higher the score, the more critical information the patch contains that contributes to the model's classification (e.g., in an image of a dog, a patch containing the dog's eyes would score higher than a patch containing the background). Then, we keep the positions of the top($N$-$k$) target patches with the highest scores fixed and replace the bottom($k$) target patches with the lowest scores with the top($k$) source patches with the highest scores. To enable the gradient backward properly to update the gate network, we opt for using Gumble Softmax~\cite{gumble} approach to achieve the selection. Hence, we can obtain the mixed embeddings $\mathcal{\Tilde{X}}^m$ of the intermediate modality which contains the key information of both source and target data.

Moreover, the parameter $k$ linearly decreases with the number of training epochs to facilitate the transition of the intermediate modality from the source to the target. For the labels of intermediate modality data, the calculation involves taking the weighted sum of the labels $y^s$ from the source data and $y^t$ target data. However, due to the disparate distributions of $y^s$ and $y^t$, this process is transformed into a weighted sum for the loss. Therefore, we can denote that the predictions of the intermediate modality data as $p^{ms}=h^s(g(\mathcal{\Tilde{X}}^m))$ for source and $p^{mt}=h^t(g(\mathcal{\Tilde{X}}^m))$ for target. The weight $\lambda$ can be calculated by $\lambda=\frac{k}{N}$. Finally, we can calculate the mixed loss $\mathcal{L}_{mix}$ using mixed embeddings $\mathcal{\Tilde{X}}^m$ as the inputs:
\begin{equation}\label{equ:loss}
  \mathcal{L}_{mix}=(1-\lambda){\mathcal{L}_{tar}}(p^{mt}, y^t)+\lambda{\mathcal{L}_{src}}(p^{ms},y^s),
\end{equation}
where $\mathcal{L}_{tar}$ is the task-specific loss for different target modalities and $\mathcal{L}_{src}$ is the CrossEntropyLoss for the source modality. The total loss of our method can be defined as: 
\begin{equation}\label{equ:loss}
  \mathcal{L}_{total}={\beta_1\mathcal{L}_{tar}}(p_t, y_t)+\beta_2\mathcal{L}_{mix},
\end{equation}
where $\beta_1$, $\beta_2$ are trade-off parameters. We summarize our \ours in Alg.~\ref{alg:pare2} in the Appendix~\ref{appendix:algorithm}.

\begin{table*}[h]
  \centering
  \caption{Prediction errors ($\downarrow$) across 10 diverse tasks on NAS-Bench-360. ``FPT" and ``NFT" respectively represent fine-tuning only the layer normalization of the model and performing one-stage full fine-tuning of the model. 
  } 
  \resizebox{\textwidth}{!}{
    \begin{tabular}{lcccccccccc}
    \toprule
    \multirow{2}[2]{*}{} & CIFAR-100 & Spherical & Darcy Flow & PSICOV & Cosmic & NinaPro & FSD50K & ECG & Satellite & DeepSEA \\
     & 0-1 error (\%) & 0-1 error (\%) & relative $\ell_2$ & MAE$_8$ & 1-AUROC & 0-1 error (\%) & 1-mAP & 1-F1 score & 0-1 error (\%) & 1-AUROC \\
    \midrule
    Hand-designed & 19.39 & 67.41 & 8.00E-03 & 3.35 & 0.127 & 8.73 & 0.62 & 0.28 & 19.80 & 0.30 \\
    \midrule
    NAS-Bench-360 & 23.39 & 48.23 & 2.60E-03 & 2.94 & 0.229 & 7.34 & 0.60 & 0.34 & 12.51 & 0.32 \\
    DASH & 24.37 & 71.28 & 7.90E-03 & 3.30 & 0.190 & 6.60 & 0.60 & 0.32 & 12.28 & 0.28 \\
    \midrule
    Perceiver IO & 70.04 & 82.57 & 2.40E-02 & 8.06 & 0.485 & 22.22 & 0.72 & 0.66 & 15.93 & 0.38 \\
    FPT & 10.11 & 76.38 & 2.10E-02 & 4.66 & 0.233 & 15.69 & 0.67 & 0.50 & 20.83 & 0.37 \\
    \midrule
    NFT & 7.67 & 55.26 & 7.34E-03 & 1.92 & 0.170 & 8.35 & 0.63 & 0.44 & 13.86 & 0.51 \\
    ORCA & 6.53 & 29.85 & 7.28E-03 & 1.91 & 0.152 & 7.54 & 0.56 & \textbf{0.28} & 11.59 & 0.29 \\
    \cellcolor{bestcolor} \ours & \cellcolor{bestcolor}\textbf{6.25} & \cellcolor{bestcolor}\textbf{25.55} & \cellcolor{bestcolor}\textbf{7.00E-03} & \cellcolor{bestcolor}\textbf{0.99} & \cellcolor{bestcolor}\textbf{0.121} & \cellcolor{bestcolor}\textbf{6.53} & \cellcolor{bestcolor}\textbf{0.55} & \cellcolor{bestcolor}\textbf{0.28} & \cellcolor{bestcolor}\textbf{11.18} & \cellcolor{bestcolor}\textbf{0.28} \\
    \bottomrule
    \end{tabular}}%
  \label{tab:nas360}%
\end{table*}%

\begin{table*}[h]
  \centering
  \caption{Normalized Root Mean Squared Errors (nRMSEs, $\downarrow$) across 8 tasks of PDEBench. \ours surpasses U-Net and PINN in all tasks, outperforms ORCA in 6 out of 8 tasks, and exhibits performance comparable to FNO.}
  \resizebox{\textwidth}{!}{
    \begin{tabular}{ccccccccccccccccc}
    \toprule
     & Advection & Burgers & Diffusion-Reaction & Diffusion-Sorption & Navier-Stokes & Darcy-Flow & Shallow-Water & Diffusion-Reaction \\
    & 1D & 1D & 1D & 1D & 1D & 2D & 2D & 2D \\
    \midrule
    PINN  & 6.70E-01 & 3.60E-01 & 6.00E-03 & 1.50E-01 & 7.20E-01 & 1.80E-01 & 8.30E-02 & 8.40E-01 \\
    FNO   & 1.10E-02 & \textbf{3.10E-03} & \textbf{1.40E-03} & 1.70E-03 & 6.80E-02 & 2.20E-01 & \textbf{4.40E-03} & \textbf{1.20E-01} \\
    U-Net & 1.10E+00 & 9.90E-01 & 8.00E-02 & 2.20E-01 & - & - & 1.70E-02 & 1.60E+00 \\
    \midrule
    ORCA  & 9.80E-03 & 1.20E-02 & 3.00E-03 & \textbf{1.60E-03} & \textbf{6.20E-02} & 8.10E-02 & 6.00E-03 & 8.20E-01 \\
    \cellcolor{bestcolor}\ours & \cellcolor{bestcolor}\textbf{2.70E-03} & \cellcolor{bestcolor}8.30E-03 & \cellcolor{bestcolor}2.60E-03 & \cellcolor{bestcolor}\textbf{1.60E-03} & \cellcolor{bestcolor}6.62E-02 & \cellcolor{bestcolor}\textbf{8.06E-02} & \cellcolor{bestcolor}5.70E-03 & \cellcolor{bestcolor}8.18E-01 \\
    \bottomrule
    \end{tabular}}%
  \label{tab:pde}%
\end{table*}%

\section{Experiments}
In this section, we first validate the effectiveness of \ours for cross-modal fine-tuning on three benchmarks: NAS-Bench-360, PDEBench and OpenML-CC18, comprising a total of 48 datasets. Subsequently, through a series of analytical experiments, we showcase the superiority of each module within \ours when compared to alternative approaches. Finally, by presenting intuitive visualization results, we illustrate the effectiveness of our gate network and the successful preservation of source knowledge in \ours.

\paragraph{Implementation details.}
We follow ORCA~\cite{orca} in using RoBERTa~\cite{roberta} and Swin Transformers~\cite{liu2021swin} as pretrained source models for 1D/2D modalities respectively, treating language and vision as the source modalities. For 2D classification tasks, CIFAR10~\cite{c10} and Tiny-ImageNet~\cite{tiny} serve as proxy datasets. For 2D dense prediction tasks, we use VOC~\cite{pascal} as a proxy dataset, modifying its labels to create a simpler foreground-background segmentation task. For 1D tasks, CoNLL-2003 is employed as a proxy dataset. For other experimental settings such as learning rates, number of epochs, optimizers, we adhere to the configurations specified by ORCA. Our experiments are conducted in a single NVIDIA RTX 4090.



\begin{figure}[h]
  \centering
  \includegraphics[width=0.4\textwidth]{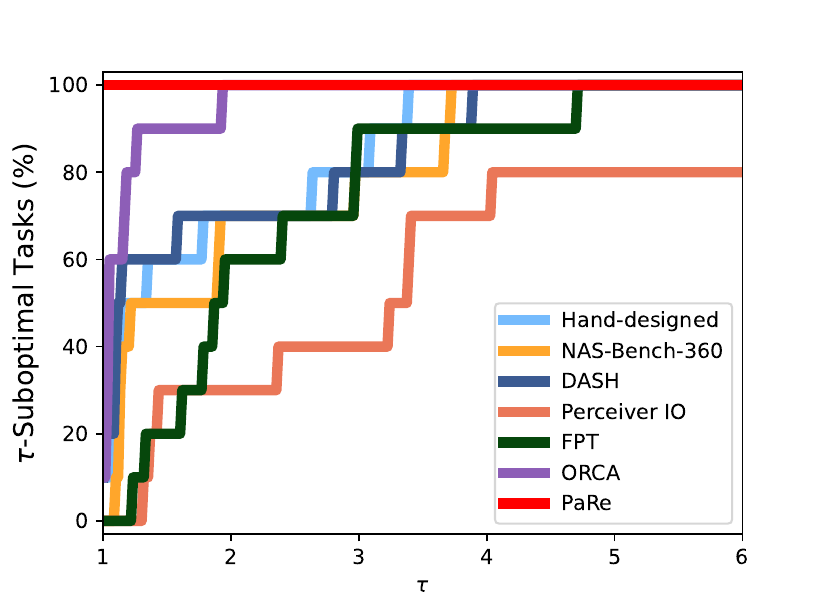}
  \caption{Aggregating Table~\ref{tab:nas360} results using performance profiles~\cite{dolan2002profiles}. The ordinate represents the cumulative distribution of problems solved by the method within a factor $\tau$ of the best performance. Therefore, the closer a curve approaches the top-left corner of the graph, the more capable the method is of solving more problems with minimal performance degradation. \ours being as a horizontal line means it is always the best.}
  \label{profile}
  \vspace{-0.8em}
\end{figure}

\begin{table}[t]
 \caption{Average classification results across 30 datasets on OpenML-CC18. ``Diff. from XGBoost" is the acrosstask average of per-task difference from XGBoost. On 15/30 datasets, \ours ranks the first among all compared methods.}
\label{table:openml}
\centering
 \Huge
\resizebox{0.48\textwidth}{!}{	\begin{tabular}{l|ccc|cc|cc}
\toprule
 OpenML-CC18 & LightGBM & CatBoost & XGBoost & AutoGluon &  TabPFN  & ORCA & \cellcolor{bestcolor}\ours \\
 \midrule
 \# Wins/Ties    &   1/30	&1/30	&2/30	&7/30	&5/30&	7/30  & \cellcolor{bestcolor}\textbf{15/30}\\
 Avg. AUROC ($\uparrow$) &  0.8840  &   0.8898  &   0.8909  &  0.8947  &  0.8943 &    0.8946 & \cellcolor{bestcolor}\textbf{0.9030}\\
 Diff. from XGBoost &  -0.0069 &   -0.0011  &   0  &  +0.0038  &    +0.0034 & +0.0036 & \cellcolor{bestcolor}\textbf{+0.0121}\\ 
\bottomrule
\end{tabular}}
 \vspace{-0.8em}
\end{table}

\begin{table*}[h]
  \centering
  \caption{Comparison prediction errors ($\downarrow$) of traditional mixing strategies and \ours variants across 10 diverse tasks, and the impact of varying strategies for different values of $k$, where ``non-gradual" indicates a constant $k$, while the other three represent different strategies for decreasing $k$.}
  \resizebox{0.95\textwidth}{!}{
    \begin{tabular}{llcccccccccc}
    \toprule
    \multicolumn{2}{c}{Method} & CIFAR-100 & Spherical & Darcy Flow & PSICOV & Cosmic & NinaPro & FSD50K & ECG & Satellite & DeepSEA \\
    \hline
    \multicolumn{2}{c}{Mixup} & 6.59 & 26.60 & 7.70E-03 & \textbf{0.99} & 0.500 & 7.74 & 0.56 & 0.29 & 11.51 & 0.29 \\
    \multicolumn{2}{c}{CutMix} & \textbf{6.11} & 27.76 & 7.20E-03 & \textbf{0.99} & 0.135 & 8.41 & 0.56 & 0.29 & 11.58 & \textbf{0.28} \\
    \midrule
    \multirow{4}{*}{\ours} 
    & w/ non-gradual & 6.59 & 27.68 & 7.20E-03 & \textbf{0.99} & 0.138 & 7.59 & 0.57 & 0.28 & 11.61 & 0.29 \\
    \cmidrule{2-12}
    & w/ piecewise & 6.22 & 26.88 & \textbf{6.90E-03} & \textbf{0.99} & 0.132 & 6.98 & 0.56 & 0.29 & \textbf{10.89} & 0.29 \\
    & w/ exponential & 6.38 & 26.35 & 7.00E-03 & \textbf{0.99} & \textbf{0.119} & 7.13 & \textbf{0.55} & \textbf{0.28} & 11.56 & \textbf{0.28} \\
    & \cellcolor{bestcolor}w/ linear (default) & \cellcolor{bestcolor}6.25 & \cellcolor{bestcolor}\textbf{25.55} & \cellcolor{bestcolor}7.00E-03 & \cellcolor{bestcolor}\textbf{0.99} & \cellcolor{bestcolor}0.121 & \cellcolor{bestcolor}\textbf{6.53} & \cellcolor{bestcolor}\textbf{0.55} & \cellcolor{bestcolor}\textbf{0.28} & \cellcolor{bestcolor}11.18 & \cellcolor{bestcolor}\textbf{0.28} \\
    \bottomrule
    \end{tabular}}%
  \label{tab:mix}
\end{table*}%

\subsection{Overall results}

\paragraph{NAS-Bench-360} comprises four 2D classification tasks, three 2D dense prediction tasks, and three 1D tasks. Here, we compare four types of baselines: (1) task-specific models designed by~\cite{nasbench360}; (2) general-purpose models exemplified by Perceiver IO~\cite{jaegle2022perceiver}; (3) AutoML methods featuring the top-performing algorithm on NAS-Bench-360, DASH~\cite{dash}; (4) cross-modal fine-tuning methods including naive fine-tuning and ORCA.

As shown in Table~\ref{tab:nas360}, \ours achieves the best performance across all tasks. Whether hand-designed or AutoML method, we demontrate significant preformance gains compares to them across multiple tasks. Particularly, in the comparison of cross-modal fine-tuning methods, we outperform ORCA on nearly all tasks. Moreover, we make substantial progress on tasks where ORCA couldn't surpass hand-designed or AutoML methods, establishing a new state-of-the-art results. 

In addition, we employ performance profiles to comprehensively compare multiple methods across a suite of datasets. Performance profiles are statistical tools used to assess and demonstrate the efficacy of optimization algorithms across a multitude of test cases. Each curve represents a different method and shows the proportion of problems it solves within varying thresholds of a performance factor $\tau$. The performance factor $\tau$ is a normalized measure of how each method's performance compares to the best performance. As shown in Fig.~\ref{profile}, \ours is always the best, which means \ours outperforms other methods across all tasks.

\paragraph{PDEBench} comprises multiple scientific ML-related datasets, with a focus on the physics domain. Following ORCA, we validate \ours on eight of these datasets. We compare our method with different SOTA task-specific models, including the physics-informed neural network PINN~\cite{PINN}, Fourier neural operator (FNO)~\cite{fno}, the generic image-to-image regression model U-Net~\cite{UNet} and ORCA~\cite{orca}. As shown in Table~\ref{tab:pde}, \ours demonstrate further improvement over ORCA across multiple datasets, achieving results to be SOTA on nearly half of the datasets among all methods.


\paragraph{OpenML-CC18 benchmark} is for tabular classification. We assess the performance of \ours across 30 datasets on OpenNL-CC18~\cite{vanschoren2014openml}. Our evaluation includes comparisons against the classical boosting methods XGBoost~\cite{Chen2016XGBoostAS}, CatBoost~\cite{Ostroumova2017CatBoostUB} and LightGBM~\cite{Ke2017LightGBMAH}, deep learning approaches like AutoGluon~\cite{Erickson2020AutoGluonTabularRA} and TabPFN~\cite{Hollmann2022TabPFNAT} and cross-modal fine-tuning method ORCA. As shown in Table~\ref{table:openml}, the average accuracy of \ours across 30 datasets is the highest among all methods, with the highest results observed in 15 out of the 30 datasets. Compared to ORCA, \ours is better on 22 datasets. The detailed results can be found in the Appendix~\ref{appendix:openml}.

In short, these results emphasize that our end-to-end strategy of constructing intermediate modalities for cross-modal fine-tuning, compared to two-stage alignment and fine-tuning of ORCA, can more efficiently close the modality gap, which enhances model transferability. In the following, we will systematically analyze the strengths of each module in \ours.

\begin{figure}[t]
    \centering
    \includegraphics[width=0.45\textwidth]{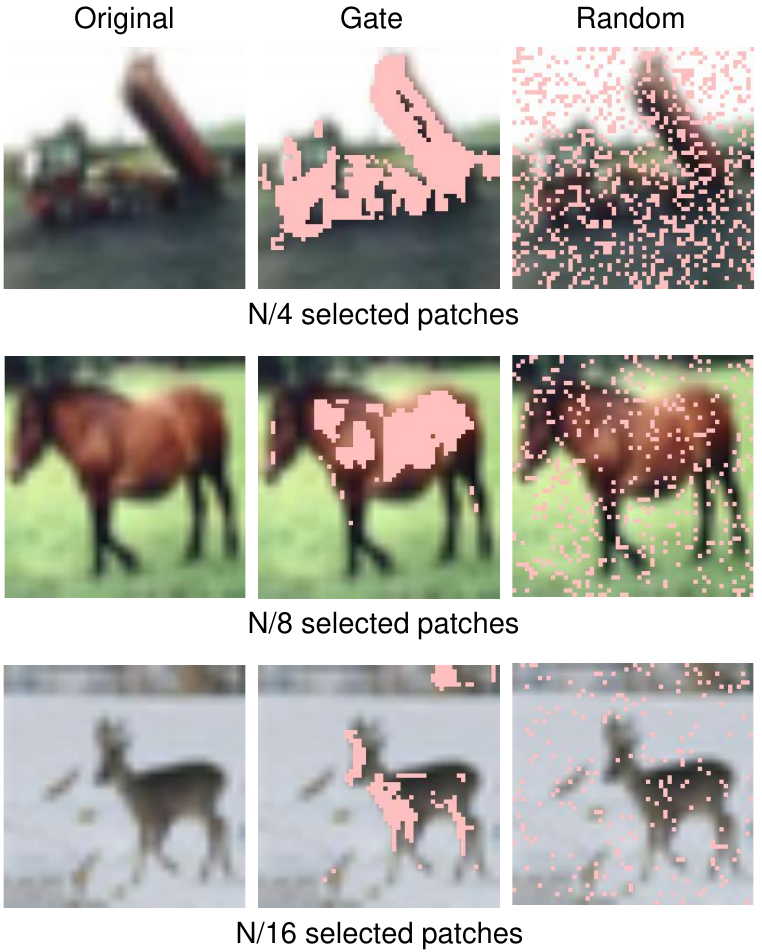}
    \caption{The visualization of the different numbers of patches selected by random strategy and our gate strategy. Additional visualizations can be found in the Appendix~\ref{appendix:pase}.}
    \label{fig:gate}
\end{figure}

\begin{table*}[t]
  \centering
  \caption{Comparison prediction errors ($\downarrow$) between different strategies (Random vs. Gate) to select patches for replacement.}
  \resizebox{0.95\textwidth}{!}{
    \begin{tabular}{lcccccccccc}
    \toprule
    & CIFAR-100 & Spherical & Darcy Flow & PSICOV & Cosmic & NinaPro & FSD50K & ECG & Satellite & DeepSEA \\
    \midrule
    Random & 6.52 & 28.06 & 7.10E-03 & \textbf{0.99} & 0.146 & 6.98 & 0.56 & 0.29 & 11.32 & \textbf{0.28} \\
    \cellcolor{bestcolor}\textbf{Gate} & \cellcolor{bestcolor}\textbf{6.25} & \cellcolor{bestcolor}\textbf{25.55} & \cellcolor{bestcolor}\textbf{7.00E-03} & \cellcolor{bestcolor}\textbf{0.99} & \cellcolor{bestcolor}\textbf{0.121} & \cellcolor{bestcolor}\textbf{6.53} & \cellcolor{bestcolor}\textbf{0.55} & \cellcolor{bestcolor}\textbf{0.28} & \cellcolor{bestcolor}\textbf{11.18} & \cellcolor{bestcolor}\textbf{0.28} \\
    \bottomrule
    \end{tabular}}%
  \label{tab:gate}%
\end{table*}%

\subsection{Why using gradually patch replacement?}
In this section, we conduct ablation studies to analyze the superiority of \ours compared to other mixing techniques. Additionally, we investigate the impact of different gradual choices for the value of $k$.

\paragraph{Comparison of the mixing ways.} There are various ways to perform mixing at the embedding space. The question is, which method can achieve modality-agnostic behavior and perform well on various target modalities.
In Table~\ref{tab:mix}, we compare our patch replacement method with patch Mixup and CutMix, which are the two most commonly used and robust methods. From the results, it is evident that our method outperforms Mixup and CutMix on almost all datasets. Particularly, on the Cosmic dataset, mixup fails to train and produces a random outcome. We attribute the superiority our method to thorough consideration of modality differences. When there is a significant modality gap, direct application of mixup way overly confuse the model, making training difficult. On the other hand, CutMix, which replaces patches in a block-wise manner, can potentially cover the critical information in the data, disrupting model training and yielding suboptimal results. Our patch replacement method ensures that the model is not overly confused. Additionally, our gating mechanism maximally preserves essential information from both modality data.

\paragraph{Comparison of different $k$-value choices.} \ours selects $k$ patches from the source embedding to replace $k$ patches in the target embedding. Next we analyze the strategy for choosing the value of $k$. We categorize the overall strategy into two types: non-gradual and gradual. For the non-gradual approach, the selection of $k$ is random, while the gradual approach involves choosing $k$ values that decrease in some form as the training progresses. As shown in Table~\ref{tab:mix}, we observe that the use of gradual strategies generally outperforms the non-gradual strategy across various datasets. The model's progression from the source to the target modality through the intermediate modality constructed by \ours, can be considered a form of curriculum learning that starts from easy and progresses to more challenging data. Thus, gradual strategies enhance the model's transferability compared to not gradual strategy. In addition, we compare different ways of decreasing the $k$ value, including piecewise, exponential and linear decreasing. We find that there may exist an optimal decreasing strategy for each task. Ultimately, \ours adopt the simplest linear decreasing strategy as our default setup across all datasets.

\subsection{How to select patches for replacement?}

When considering patch selection from source embeddings to replace patches in target embeddings, a straightforward approach is to randomly choose patches for replacement. While random replacement indeed leads to the generation of intermediate modalities, as illustrated in Fig.~\ref{fig:gate}, the right column represents patches selected through random patch selection, highlighted for visualization. From this column, we observe that due to the randomness of replacement, there is a possibility to select too many background patches. Additionally, the critical patches are not continuous, which might hinder the model from capturing essential information in the data and, instead, disrupt the model's training.

\begin{figure}[htbp]
    \centering
    \includegraphics[width=0.4\textwidth]{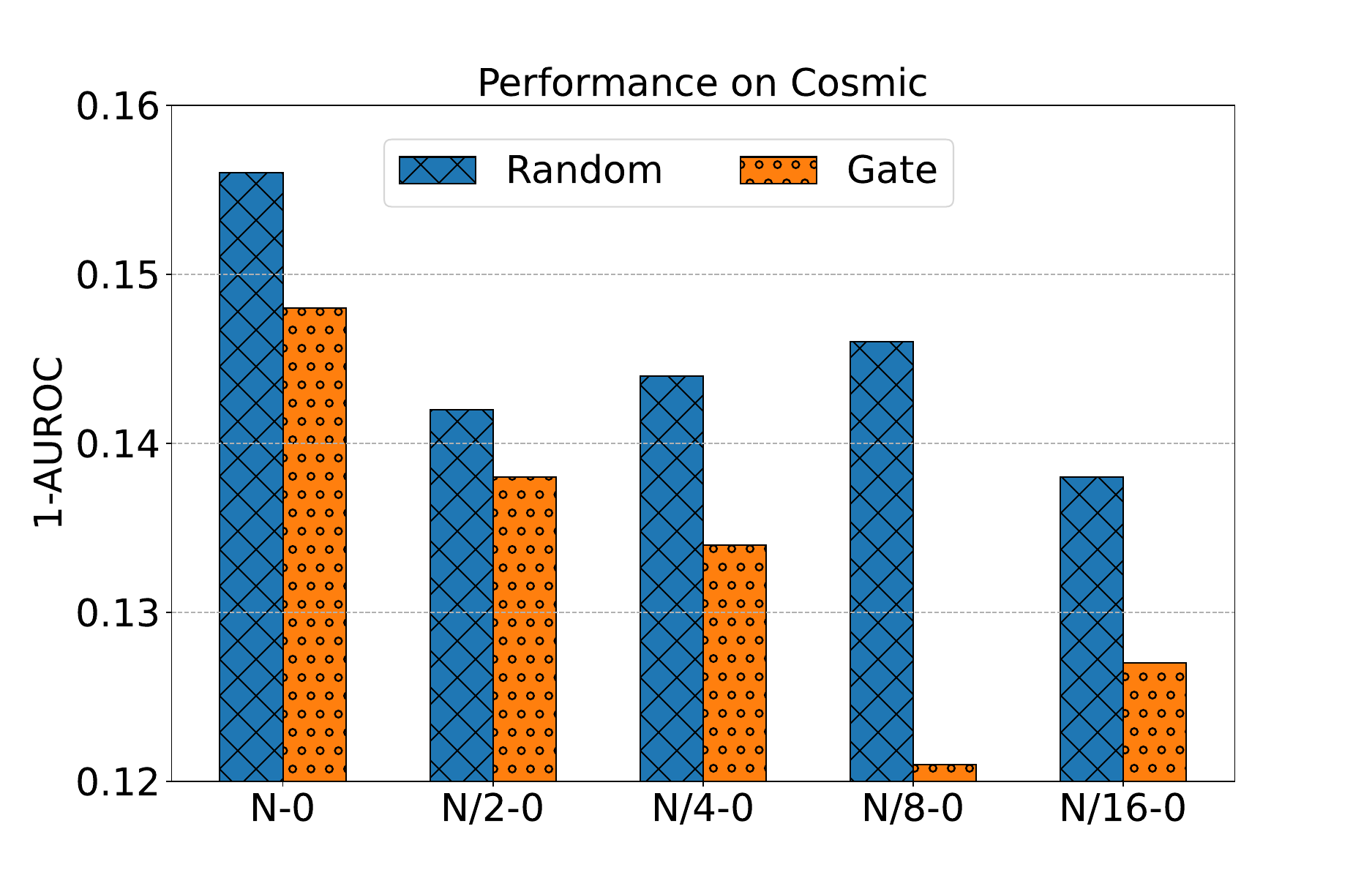}
    \caption{The impact on the results of random and gate strategy on Cosmic dataset with different initial $k$ value. The smaller the initial $k$ value, the larger performance percentage difference between random and gate strategy.}
    \label{fig:fig_cosmic}
    \vspace{-0.8em}
\end{figure}

Therefore, we employ a gate network for patch scoring. Based on the scores, we select the top-$k$ scored patches from the source to replace the bottom-$k$ scored patches in the target. This approach allows us to preserve critical information from both source and target data as much as possible. The middle column in Fig.~\ref{fig:gate} represents patches selected through patch scoring. We observe that patches chosen through this strategy focus more on the main parts of the data, enabling more thorough training of the model on the intermediate modality.

In Table~\ref{tab:gate}, we present a comparison between the results of selecting patches randomly and using the gate network for patch selection. We observe that, across all datasets, the approach using the gate network consistently outperforms the random selection method. This demonstrates the effectiveness of the gate network in patch scoring. Furthermore, in Fig.~\ref{fig:fig_cosmic}, we compare the results on the Cosmic dataset when choosing different initial values for $k$ and gradually decreasing it to 0 during the training process for both random and patch scoring methods. From the results, we observe that as the initial $k$ value decreases (e.g., N-0 and N/8-0), the performance gap between the patch scoring method and the random method widens. This suggests that the advantage of our gate network becomes more pronounced when a smaller number of patches are selected.

To conduct a more comprehensive analysis, we configure the gate network with an fully connected (FC) layer followed by a sigmoid function, an MLP followed by a sigmoid function and an MLP followed by Dropout and a sigmoid function. In Table~\ref{tab:gateconfig}, we found that increasing the complexity of the gate network may not necessarily lead to improvement. This could be due to the limited amount of target modality data, and the increased complexity of the network may result in insufficient training. Therefore, we use the simplest gate network structure (FC layer followed by a sigmoid function) as the version of our method.

\begin{table*}[t]
  \centering
  \caption{Comparison prediction errors ($\downarrow$) between different configuration of gate network.}
  \vspace{-0.5em}
  \resizebox{0.95\textwidth}{!}{
    \begin{tabular}{ccccccccccccccccc}
    \toprule
    \multicolumn{3}{c}{}  & \multicolumn{2}{c}{CIFAR-100} & \multicolumn{2}{c}{Spherical} & \multicolumn{2}{c}{Darcy Flow} & \multicolumn{2}{c}{PSICOV} & \multicolumn{2}{c}{Cosmic} & \multicolumn{2}{c}{NinaPro} & \multicolumn{2}{c}{FSD50K} \\
    \midrule
    \multicolumn{3}{c}{\cellcolor{bestcolor}FC + sigmoid (\textbf{Ours})} & \multicolumn{2}{c}{\cellcolor{bestcolor}6.25} & \multicolumn{2}{c}{\cellcolor{bestcolor}\textbf{25.55}} & \multicolumn{2}{c}{\cellcolor{bestcolor}7.00E-03} & \multicolumn{2}{c}{\cellcolor{bestcolor}\textbf{0.99}} & \multicolumn{2}{c}{\cellcolor{bestcolor}\textbf{0.121}} & \multicolumn{2}{c}{\cellcolor{bestcolor}\textbf{6.53}} & \multicolumn{2}{c}{\cellcolor{bestcolor}\textbf{0.55}} \\
    \multicolumn{3}{c}{MLP + sigmoid} & \multicolumn{2}{c}{6.40} & \multicolumn{2}{c}{26.08} & \multicolumn{2}{c}{\textbf{6.90E-03}} & \multicolumn{2}{c}{\textbf{0.99}} & \multicolumn{2}{c}{0.125} & \multicolumn{2}{c}{6.74} & \multicolumn{2}{c}{\textbf{0.55}} \\
    \multicolumn{3}{c}{MLP + dropout + sigmoid} & \multicolumn{2}{c}{\textbf{6.24}} & \multicolumn{2}{c}{26.52} & \multicolumn{2}{c}{\textbf{6.90E-03}} & \multicolumn{2}{c}{\textbf{0.99}} & \multicolumn{2}{c}{0.131} & \multicolumn{2}{c}{6.74} & \multicolumn{2}{c}{\textbf{0.55}} \\
    \bottomrule
    \end{tabular}}%
  \label{tab:gateconfig}%
  \vspace{-0.5em}
\end{table*}%

\subsection{Influence of proxy source datasets} \label{srcdataset}

By deploying the patch replacement algorithm in the embedding space, \ours facilitates the alignment of modalities, allowing for the utilization of a wide range of proxy source datasets. This promotes generalization across intermediate modalities, thus enhancing the model's performance. To demonstrate that the primary advancement of \ours lies in its ability to gradually bridge the modality gap, rather than simply resulting from the substitution of different proxy source datasets. We conduct an ablation study in Table~\ref{tab:prosrc} using different proxy source datasets. Specifically, we utilize CIFAR-10, Caltech101 and Tiny-ImageNet as the proxy source datasets for the 2D classification task. As shown in Table~\ref{tab:prosrc}, significant improvements of \ours is over ORCA~\cite{orca} across all scenarios, indicating that the efficacy of \ours stems from its ability to bridge the modality gap rather than simply swapping proxy source datasets. Besides, \ours achieves good results with multiple proxy source datasets with relatively robust performance.

\begin{table}[htbp]
  \centering
  \Huge
  \caption{The influence of using CIFAR-10, Caltech101 and Tiny-ImageNet as the proxy source datasets for 2D classification task.}
  \resizebox{0.48\textwidth}{!}{
    \begin{tabular}{lccccccccccc}
    \toprule
    \multicolumn{4}{c}{\textbf{2D Classification}} & \multicolumn{2}{c}{CIFAR-100} & \multicolumn{2}{c}{Spherical} & \multicolumn{2}{c}{NinaPro} & \multicolumn{2}{c}{FSD50K} \\
    \midrule
    Method & \multicolumn{3}{c}{proxy source dataset} & \multicolumn{2}{c}{0-1error(\%)} & \multicolumn{2}{c}{0-1error(\%)} & \multicolumn{2}{c}{0-1error(\%)} & \multicolumn{2}{c}{1-mAP} \\
    \midrule
    ORCA  & \multicolumn{3}{c}{\multirow{2}[2]{*}{CIFAR10}} & \multicolumn{2}{c}{6.53} & \multicolumn{2}{c}{29.85} & \multicolumn{2}{c}{7.54} & \multicolumn{2}{c}{0.56} \\
    \cellcolor{bestcolor}\textbf{PaRe} & \multicolumn{3}{c}{}  & \multicolumn{2}{c}{\cellcolor{bestcolor}\textbf{6.25}} & \multicolumn{2}{c}{\cellcolor{bestcolor}\textbf{26.47}} & \multicolumn{2}{c}{\cellcolor{bestcolor}\textbf{6.53}} & \multicolumn{2}{c}{\cellcolor{bestcolor}\textbf{0.55}} \\
    \midrule
    ORCA  & \multicolumn{3}{c}{\multirow{2}[2]{*}{Caltech101}} & \multicolumn{2}{c}{6.48} & \multicolumn{2}{c}{29.64} & \multicolumn{2}{c}{8.04} & \multicolumn{2}{c}{0.57} \\
    \cellcolor{bestcolor}\textbf{PaRe} & \multicolumn{3}{c}{}  & \multicolumn{2}{c}{\cellcolor{bestcolor}\textbf{6.39}} & \multicolumn{2}{c}{\cellcolor{bestcolor}\textbf{26.22}} & \multicolumn{2}{c}{\cellcolor{bestcolor}\textbf{7.19}} & \multicolumn{2}{c}{\cellcolor{bestcolor}\textbf{0.56}} \\
    \midrule
    ORCA  & \multicolumn{3}{c}{\multirow{2}[2]{*}{Tiny-ImageNet}} & \multicolumn{2}{c}{6.44} & \multicolumn{2}{c}{28.21} & \multicolumn{2}{c}{8.35} & \multicolumn{2}{c}{0.56} \\
    \cellcolor{bestcolor}\textbf{PaRe} & \multicolumn{3}{c}{}  & \multicolumn{2}{c}{\cellcolor{bestcolor}\textbf{6.25}} & \multicolumn{2}{c}{\cellcolor{bestcolor}\textbf{25.55}} & \multicolumn{2}{c}{\cellcolor{bestcolor}\textbf{7.59}} & \multicolumn{2}{c}{\cellcolor{bestcolor}\textbf{0.55}} \\
    \bottomrule
    \end{tabular}}%
  \label{tab:prosrc}%
  \vspace{-0.5em}
\end{table}%

\subsection{Visualization of source knowledge preservation} \label{srcknow}

\begin{figure}
    \centering
    \includegraphics[width=0.46\textwidth]{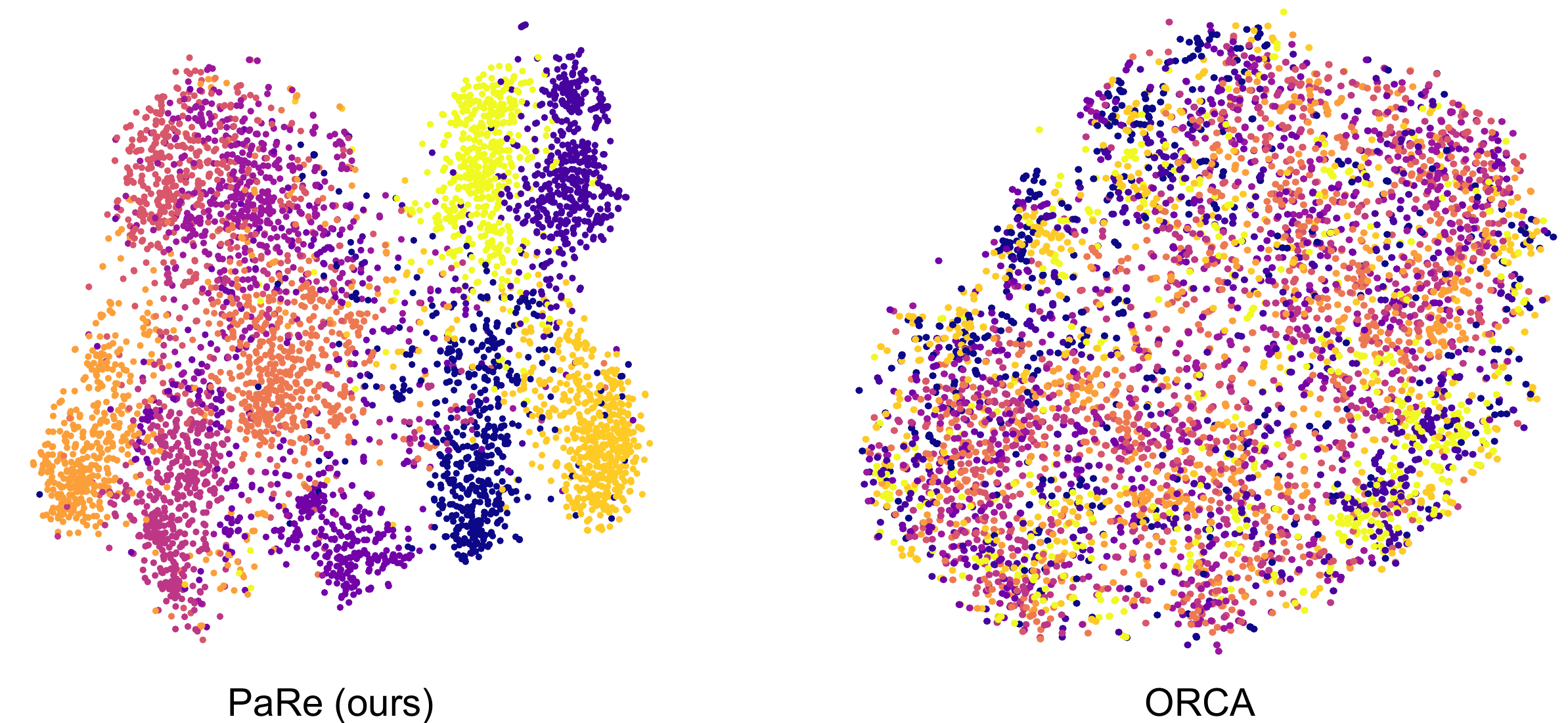}
    \caption{The t-SNE of the feature of \ours and ORCA on source proxy dataset CIFAR10, which indicates the ability of source knowledge preservation.}
    \label{fig:ana}
    \vspace{-1.5em}
\end{figure}

The goal of cross-modal fine-tuning is to adapt a pretrained model with rich knowledge in the source modality to a specific target modality. Intuitively, if the pretrained model can effectively retain the knowledge from the source modality during cross-modal fine-tuning, it can better leverage the abundant knowledge for transfer. Therefore, we obtain two models fine-tuned on Ninapro with \ours and ORCA separately, and we compare the t-SNE visualization of the CIFAR10 features they output in Fig.~\ref{fig:ana}.

As shown in Fig.~\ref{fig:ana}, although ORCA narrows the OTDD between source and target modality data in the first stage, it still fails to preserve source modality knowledge due to significant modality gap. In contrast, \ours utilizes intermediate modality data for training, requiring accurate feature extraction from source patches for correct classification. Therefore, knowledge from source modality can be well-preserved, leading to better transferability of the model.


\section{Conclusion}
In this paper, we propose an end-to-end cross-modal fine-tuning method, \ours, employing patch scoring for patch replacement between the source and target modality data. \ours facilitates the generation of intermediate modalities that progress from easy to hard during training, bridging the modality gap to enhance training stability and model tranferability, while also mitigating the challenge of limited data in the target modality. \ours achieve good performance on three benchmarks consisting of 48 datasets, presenting a novel transfer methodology for cross-modal fine-tuning.

\textbf{Limitation and furture work.}
During the course of this research, we identify some limitations in the current version and directions for future improvement: Firstly, since our method does not require training with computationally expensive OTDD~\cite{otdd}, our approach exhibits high scalability with respect to the source modality proxy dataset. However, determining the most suitable source modality proxy dataset based on the target modality dataset remains a challenge. Secondly, we observe that solely augmenting the target modality data can yield better results in certain modalities. However, these data augmentation methods are not universal. Therefore, a modality-agnostic data augmentation method is necessary to prevent model overfitting and enhance cross-modal fine-tuning. Lastly, leveraging unlabeled data from the target modality is also a promising direction for research, as it can better alleviate the issue of insufficient target modality data.

\section*{Acknowledgements}

This paper was supported by the National Natural Science Foundation of China (No. 62376026), Beijing Nova Program (No. 20230484296) and CCF-Tencent Rhino-Bird Open Research Fund.


\section*{Impact Statement}

The application of cross-modal transfer learning in this paper demonstrates significant performance improvements by transferring pretrained vision or language models to other modalities such as PDEs, protein structures, cosmic rays, gestures, and so on. This approach can drive advancements in diverse fields, promote interdisciplinary research, and enhance the robustness and versatility of models. By improving efficiency and accessibility, our work has the potential to foster novel discoveries, improve human-computer interaction, and contribute to societal and economic benefits.






\bibliography{example_paper}
\bibliographystyle{icml2024}

\newpage
\appendix
\onecolumn
\newpage
\appendix
\section{Appendix}

\subsection{Algorithm of \ours} \label{appendix:algorithm}

The algorithm of \ours is illustrated in Alg.~\ref{alg:pare2}. Firstly, we score the patches from both the source and target using a gate network to obtain scores for each patch. Next, we select the top-$k$ scored patches to replace the bottom-$k$ scored target patches one by one. To ensure the backpropagation of gradients, we employ Gumble softmax operation to select the top-$k$ and bottom-$k$ patches. Finally, we update the model by separately calculating the source and target loss.

\begin{algorithm}[htbp]
\caption{\small{Pseudocode of \ours.}}
\label{alg:pare2}

\definecolor{codeblue}{rgb}{0.25,0.5,0.5}
\lstset{
  backgroundcolor=\color{white},
  basicstyle=\fontsize{7.2pt}{7.2pt}\ttfamily\selectfont,
  columns=fullflexible,
  breaklines=true,
  captionpos=b,
  commentstyle=\fontsize{7.2pt}{7.2pt}\color{codeblue},
  keywordstyle=\fontsize{7.2pt}{7.2pt},
}
\begin{center}
\begin{lstlisting}[language=python]
# N: the embeddings length
# x_t, x_s: target and source embeddings 
# y_t, y_s: target and source labels
# model.gate: the gate netowrk of the model
# SubsetOperator: using gumble softmax to do selection

k = int(N - N * (current_epoch / totle_epoch))
for (x_t, y_t, x_s, y_s) in loader:
    # Patch Scoring using Gate Network
    tar_score = model.gate(x_t)
    src_score = model.gate(x_s)
    
    # select bottom-k and top-k
    bk_tar_mask = 1 - SubsetOperator(tar_score, N-k)
    tk_src_mask = SubsetOperator(src_score, k)
    bk_tar_indices = torch.nonzero(bk_tar_mask)
    tk_tar_indices = torch.nonzero(tk_tar_mask)
    
    # Patch Replacement
    x_mix = x_t.clone()
    x_mix = x_mix * (1 - bk_tar_mask)
    x_s = x_s * tk_src_mask
    x_mix[bk_tar_indices] = x_s[tk_src_indices]
    
    # Obtain Logits
    logits_mix_tar, logits_mix_src = model(x_mix)
    
    # Calculate Loss
    lam = k / N
    L_tar = L_tar(logits_mix_tar, y_t)
    L_src = L_src(logits_mix_src, y_s)
    L_mix = (1-lam) * L_tar + lam * L_src
\end{lstlisting}
\end{center}
\end{algorithm}

\subsection{Implementation details} \label{appendix:implementation}
\subsubsection{Pretrained Models}
We evaluate \ours with two pretrained models in our experiments. For all 2D tasks we use Swin-base~\cite{liu2021swin} pretrained on ImageNet-22K, and for all 1D tasks, we use RoBERTa-base~\cite{roberta} pretrained on Five English-language corpora. We follow ORCA~\cite{orca} use the Hugging Face transformers library~\cite{wolf2019huggingface} to implement the pretrained models.
\subsubsection{NAS-Bench-360}

NAS-Bench-360 encompasses 10 tasks across various modalities, such as image classification, hand gesture recognition, and solving PDEs, among others. These tasks have diverse objectives, including 2D classification, 2D dense prediction, and 1D classification. They involve different types of data and are trained using distinct loss functions. Additionally, each task is associated with its unique hand-designed expert model. The Table~\ref{tab:nasintro} provides an overview of NAS-Bench-360.

\begin{table*}[htbp]
  \centering
  \LARGE
  \caption{The introduction to the 10 tasks in NAS-Bench-360.}
  \resizebox{0.98\textwidth}{!}{
    \begin{tabular}{ccccccl}
    \toprule
    Dateset & DATA\_NUM & DATA\_DIM & TYPE & CLASS\_NUM & Loss & Expert arch. \\
    \midrule
    CIFAR100 & 60K & 2D & Point & 100 & CE & DenseNet-BC~\cite{Huang2017DenselyCC} \\
    Spherical & 60K & 2D & Point & 100 & CE & S2CN~\cite{Cohen2018SphericalC} \\
    NinaPro & 3,956 & 2D & Point & 18 & LpLoss & Attention Model~\cite{Josephs2020sEMGGR} \\
    FSD50K & 51K & 2D & Point & 200 & MSELoss & VGG~\cite{Fonseca2021FSD50KAO} \\
    Darcy Flow & 1.1K & 2D & Dense & 1 & BCE & FNO~\cite{fno} \\
    PSICOV & 3,606 & 2D & Dense & 1 & FocalLoss & DEEPCON~\cite{deepcon} \\
    Cosmic & 5,250 & 2D & Dense & 1 & BCE & deepCR-mask~\cite{Zhang2019deepCRCR} \\
    ECG & 330K & 1D & Point & 4 & CE & ResNet-1D~\cite{Hong2020HOLMESHO} \\
    Satellite & 1M & 1D & Point & 24 & CE & ROCKET~\cite{Dempster2020ROCKETEF} \\
    DeepSEA & 250K & 1D & Point & 36 & BCE & DeepSEA~\cite{Zhou2015PredictingEO} \\
    \bottomrule
    \end{tabular}
    }
  \label{tab:nasintro}%
\end{table*}%

\subsubsection{PDEBench}

PDEBench comprises multiple datasets of partial differential equations (PDEs) with varying parameters and initial conditions. Following ORCA, we utilize the eight datasets listed in the Table~\ref{tab:pdeintro}, which provides detailed information on dataset parameters, utilized loss functions, and other relevant details.

\begin{table*}[htbp]
  \centering
  \LARGE
  \caption{The introduction to the 8 tasks in PDEBench.}
  \resizebox{0.95\textwidth}{!}{
    \begin{tabular}{cccccccccccccccccc}
    \toprule
    \multicolumn{2}{c}{} & \multicolumn{2}{c}{Advection} & \multicolumn{2}{c}{Burgers} & \multicolumn{2}{c}{Diffusion-Reaction} & \multicolumn{2}{c}{Diffusion-Sorption} & \multicolumn{2}{c}{Navier-Stokes} & \multicolumn{2}{c}{Darcy-Flow} & \multicolumn{2}{c}{Shallow-Water} & \multicolumn{2}{c}{Diffusion-Reaction} \\
    \midrule
    \multicolumn{2}{c}{DATA\_DIM} & \multicolumn{2}{c}{1D} & \multicolumn{2}{c}{1D} & \multicolumn{2}{c}{1D} & \multicolumn{2}{c}{1D} & \multicolumn{2}{c}{1D} & \multicolumn{2}{c}{2D} & \multicolumn{2}{c}{2D} & \multicolumn{2}{c}{2D} \\
    \midrule
    \multicolumn{2}{c}{TYPE} & \multicolumn{16}{c}{Dense Prediction} \\
    \midrule
    \multicolumn{2}{c}{Resolution} & \multicolumn{2}{c}{1024} & \multicolumn{2}{c}{1024} & \multicolumn{2}{c}{1024} & \multicolumn{2}{c}{1024} & \multicolumn{2}{c}{1024} & \multicolumn{2}{c}{128$\times$128} & \multicolumn{2}{c}{128$\times$128} & \multicolumn{2}{c}{128$\times$128} \\
    \midrule
    \multicolumn{2}{c}{Parameters} & \multicolumn{2}{c}{$\beta=0.4$} & \multicolumn{2}{c}{$\nu=1.0$} & \multicolumn{2}{c}{$\nu=0.5$, $\rho=1.0$} & \multicolumn{2}{c}{-} & \multicolumn{2}{c}{$\eta = \zeta=0.1$} & \multicolumn{2}{c}{$\beta=0.1$} & \multicolumn{2}{c}{-} & \multicolumn{2}{c}{-} \\
    \midrule
    \multicolumn{2}{c}{Loss} & \multicolumn{16}{c}{Normalized Root Mean Squared Errors (nRMSEs)} \\
    \bottomrule
    \end{tabular}}%
  \label{tab:pdeintro}%
\end{table*}%

\subsubsection{OpenML-CC18 Benchmark}

We follow ORCA to evaluate \ours on 30 datasets as shown in Table~\ref{tab:openmlintro} from OpenML-CC18 benchmark, and we follow TabPFN to use the same evaluation protocol and use the one-vs-one AUROC as the score metric. The train-test split ratio is 0.5:0.5 to account for the limited context length of TabPFN. As for training, we employ the cross-entropy loss, with the class weights set to $1/(num\_of\_samples)$.

\begin{table}[htbp]
  \centering
  \caption{The introduction of the 30 datasets in OpenML-CC18 benchmark.}
    \begin{tabular}{cccccc|cccccc}
    \toprule
    \multicolumn{2}{c}{OpenML ID} & \multicolumn{3}{c}{Name} & \#Class. & \multicolumn{2}{c}{OpenML ID} & \multicolumn{3}{c}{Name} & \#Class. \\
    \midrule
    \multicolumn{2}{c}{11} & \multicolumn{3}{c}{balance-scale} & 3     & \multicolumn{2}{c}{1049} & \multicolumn{3}{c}{pc4} & 2 \\
    \multicolumn{2}{c}{14} & \multicolumn{3}{c}{mfeat-fourier} & 10    & \multicolumn{2}{c}{1050} & \multicolumn{3}{c}{pc3} & 2 \\
    \multicolumn{2}{c}{15} & \multicolumn{3}{c}{breast-w} & 2     & \multicolumn{2}{c}{1063} & \multicolumn{3}{c}{kc2} & 2 \\
    \multicolumn{2}{c}{16} & \multicolumn{3}{c}{mfeat-karhunen} & 10    & \multicolumn{2}{c}{1068} & \multicolumn{3}{c}{pc1} & 2 \\
    \multicolumn{2}{c}{18} & \multicolumn{3}{c}{mfeat-morphological} & 10    & \multicolumn{2}{c}{1462} & \multicolumn{3}{c}{banknote-authenti…} & 2 \\
    \multicolumn{2}{c}{22} & \multicolumn{3}{c}{mfeat-zernike} & 10    & \multicolumn{2}{c}{1464} & \multicolumn{3}{c}{blood-transfusion-…} & 2 \\
    \multicolumn{2}{c}{23} & \multicolumn{3}{c}{cmc} & 3     & \multicolumn{2}{c}{1480} & \multicolumn{3}{c}{ilpd} & 2 \\
    \multicolumn{2}{c}{29} & \multicolumn{3}{c}{credit-approval} & 2     & \multicolumn{2}{c}{1494} & \multicolumn{3}{c}{qsar-biodeg} & 2 \\
    \multicolumn{2}{c}{31} & \multicolumn{3}{c}{credit-g} & 2     & \multicolumn{2}{c}{1510} & \multicolumn{3}{c}{wdbc} & 2 \\
    \multicolumn{2}{c}{37} & \multicolumn{3}{c}{diabetes} & 2     & \multicolumn{2}{c}{6332} & \multicolumn{3}{c}{cylinder-bands} & 2 \\
    \multicolumn{2}{c}{50} & \multicolumn{3}{c}{tic-tac-toe} & 2     & \multicolumn{2}{c}{23381} & \multicolumn{3}{c}{dresses-sales} & 2 \\
    \multicolumn{2}{c}{54} & \multicolumn{3}{c}{vehicle} & 4     & \multicolumn{2}{c}{40966} & \multicolumn{3}{c}{MiceProtein} & 8 \\
    \multicolumn{2}{c}{188} & \multicolumn{3}{c}{eucalyptus} & 5     & \multicolumn{2}{c}{40975} & \multicolumn{3}{c}{car} & 4 \\
    \multicolumn{2}{c}{458} & \multicolumn{3}{c}{analcatdata auth…} & 4     & \multicolumn{2}{c}{40982} & \multicolumn{3}{c}{steel-plates-fault} & 7 \\
    \multicolumn{2}{c}{469} & \multicolumn{3}{c}{analcatdata dmft} & 6     & \multicolumn{2}{c}{40994} & \multicolumn{3}{c}{climate-model-simu…} & 2 \\
    \bottomrule
    \end{tabular}%
  \label{tab:openmlintro}%
\end{table}%

\subsubsection{Proxy Source Datasets}
Although we don't need to follow the same approach as ORCA, which involves reducing the distance between the target and source modality at the embedding level in the first stage, we still require a proxy dataset for the source modality as we construct intermediate modalities through source and target data during end-to-end cross-modal fine-tuning. It's worth mentioning that, since we don't need to calculate the loss using the computationally expensive OTDD, our source proxy dataset is more scalable compared to ORCA, with a relatively minor increase in computational complexity when using additional samples. Therefore, we use CIFAR10 and Tiny-imagenet as proxy datasets for 2D classification tasks, reset the labels of PASCAL VOC to 0 and 1, effectively treating it as a foreground-background segmentation task for 2D dense prediction tasks, and employ CONLL-2003 as the proxy dataset for both 1D classification and dense prediction tasks.

Our experiments in Tabel~\ref{tab:proxy} revealed that choosing different source proxy datasets may have varied effects on different modalities. Therefore, determining how to select appropriate source proxy datasets based on different modalities is also a direction for future research in our method.

\begin{table*}[htbp]
  \centering
  \caption{Varied effects on different modalities while choosing different source proxy datasets.}
    \begin{tabular}{cccccccccc}
    \toprule
    \multicolumn{2}{c}{Proxy Dataset} & \multicolumn{2}{c}{num\_sample} & \multicolumn{2}{c}{Spherical} & \multicolumn{2}{c}{Darcy Flow} & \multicolumn{2}{c}{Ninapro} \\
    \midrule
    \multicolumn{2}{c}{\multirow{2}[2]{*}{CIFAR10}} & \multicolumn{2}{c}{5000} & \multicolumn{2}{c}{26.69} & \multicolumn{2}{c}{7.70E-03} & \multicolumn{2}{c}{7.13} \\
    \multicolumn{2}{c}{} & \multicolumn{2}{c}{all} & \multicolumn{2}{c}{26.47} & \multicolumn{2}{c}{7.90E-03} & \multicolumn{2}{c}{\textbf{6.53}} \\
    \midrule
    \multicolumn{2}{c}{\multirow{2}[2]{*}{Tiny\_ImageNet}} & \multicolumn{2}{c}{5000} & \multicolumn{2}{c}{27.18} & \multicolumn{2}{c}{-} & \multicolumn{2}{c}{7.44} \\
    \multicolumn{2}{c}{} & \multicolumn{2}{c}{all} & \multicolumn{2}{c}{\textbf{25.55}} & \multicolumn{2}{c}{-} & \multicolumn{2}{c}{7.59} \\
    \midrule
    \multicolumn{2}{c}{PASCAL\_VOC} & \multicolumn{2}{c}{all} & \multicolumn{2}{c}{-} & \multicolumn{2}{c}{\textbf{7.00E-03}} & \multicolumn{2}{c}{-} \\
    \bottomrule
    \end{tabular}%
  \label{tab:proxy}%
\end{table*}%

\subsubsection{Hyperparameters}
Due to the multitude of tasks across different modalities, it's challenging to define a single set of fine-tuning hyperparameters for all models or tasks. Therefore, we adopt the exact same hyperparameters as ORCA~\cite{orca} for model fine-tuning to facilitate comparison. The specific parameter settings are shown in the Tabel~\ref{tab:nashyper} and Table~\ref{tab:pdehyper}. For our method's hyperparameters, one concerns the initial and final values of $k$, while the other relates to the loss trade-off $\beta_1$ and $\beta_2$.

For the setting of $k$, we set the initial value to 3000 and the final value to 0 for all tasks except for the Cosmic dataset. Since Cosmic involves a binary classification dense prediction task, excessively large $k$ values may overly interfere with the model. Therefore, for Cosmic, we set the initial value to 200 and the final value to 0. For all tasks, we uniformly set both $\beta_1$ and $\beta_2$ to 1.0.

\begin{table*}[htbp]
  \centering
  \LARGE
  \caption{The training hyperparameters of the 10 tasks on NAS-Bench-360.}
  \resizebox{0.95\textwidth}{!}{
    \begin{tabular}{cccccccccccccccccccccc}
    \toprule
    \multicolumn{2}{c}{} & \multicolumn{2}{c}{CIFAR100} & \multicolumn{2}{c}{Spherical} & \multicolumn{2}{c}{NinaPro} & \multicolumn{2}{c}{FSD50K} & \multicolumn{2}{c}{Darcy Flow} & \multicolumn{2}{c}{PSICOV} & \multicolumn{2}{c}{Cosmic} & \multicolumn{2}{c}{ECG} & \multicolumn{2}{c}{Satellite} & \multicolumn{2}{c}{DeepSEA} \\
    \midrule
    \multicolumn{2}{c}{Batch Size} & \multicolumn{2}{c}{32} & \multicolumn{2}{c}{32} & \multicolumn{2}{c}{32} & \multicolumn{2}{c}{32} & \multicolumn{2}{c}{4} & \multicolumn{2}{c}{1} & \multicolumn{2}{c}{4} & \multicolumn{2}{c}{4} & \multicolumn{2}{c}{16} & \multicolumn{2}{c}{16} \\
    \midrule
    \multicolumn{2}{c}{Epoch} & \multicolumn{2}{c}{60} & \multicolumn{2}{c}{60} & \multicolumn{2}{c}{60} & \multicolumn{2}{c}{100} & \multicolumn{2}{c}{100} & \multicolumn{2}{c}{10} & \multicolumn{2}{c}{60} & \multicolumn{2}{c}{15} & \multicolumn{2}{c}{60} & \multicolumn{2}{c}{13} \\
    \midrule
    \multicolumn{2}{c}{Accum.} & \multicolumn{2}{c}{32} & \multicolumn{2}{c}{4} & \multicolumn{2}{c}{1} & \multicolumn{2}{c}{1} & \multicolumn{2}{c}{1} & \multicolumn{2}{c}{32} & \multicolumn{2}{c}{1} & \multicolumn{2}{c}{16} & \multicolumn{2}{c}{4} & \multicolumn{2}{c}{1} \\
    \midrule
    \multicolumn{2}{c}{Optimizer} & \multicolumn{2}{c}{SGD} & \multicolumn{2}{c}{AdamW} & \multicolumn{2}{c}{Adam} & \multicolumn{2}{c}{Adam} & \multicolumn{2}{c}{AdamW} & \multicolumn{2}{c}{Adam} & \multicolumn{2}{c}{AdamW} & \multicolumn{2}{c}{SGD} & \multicolumn{2}{c}{AdamW} & \multicolumn{2}{c}{Adam} \\
    \midrule
    \multicolumn{2}{c}{Learning Rate} & \multicolumn{2}{c}{1.00E-04} & \multicolumn{2}{c}{1.00E-04} & \multicolumn{2}{c}{1.00E-04} & \multicolumn{2}{c}{1.00E-04} & \multicolumn{2}{c}{1.00E-03} & \multicolumn{2}{c}{5.00E-06} & \multicolumn{2}{c}{1.00E-03} & \multicolumn{2}{c}{1.00E-06} & \multicolumn{2}{c}{3.00E-05} & \multicolumn{2}{c}{1.00E-05} \\
    \midrule
    \multicolumn{2}{c}{Weight Decay} & \multicolumn{2}{c}{1.00E-03} & \multicolumn{2}{c}{1.00E-01} & \multicolumn{2}{c}{1.00E-05} & \multicolumn{2}{c}{5.00E-05} & \multicolumn{2}{c}{5.00E-03} & \multicolumn{2}{c}{1.00E-05} & \multicolumn{2}{c}{0.00E+00} & \multicolumn{2}{c}{1.00E-01} & \multicolumn{2}{c}{3.00E-06} & \multicolumn{2}{c}{0.00E+00} \\
    \bottomrule
    \end{tabular}}%
  \label{tab:nashyper}%
\end{table*}%

\begin{table*}[htbp]
  \centering
  \LARGE
  \caption{The training hyperparameters of the 8 tasks on PDEBench.}
  \resizebox{0.95\textwidth}{!}{
    \begin{tabular}{cccccccccccccccccc}
    \toprule
    \multicolumn{2}{c}{} & \multicolumn{2}{c}{Advection} & \multicolumn{2}{c}{Burgers} & \multicolumn{2}{c}{Diffusion-Reaction} & \multicolumn{2}{c}{Diffusion-Sorption} & \multicolumn{2}{c}{Navier-Stokes} & \multicolumn{2}{c}{Darcy-Flow} & \multicolumn{2}{c}{Shallow-Water} & \multicolumn{2}{c}{Diffusion-Reaction} \\
    \midrule
    \multicolumn{2}{c}{Batch Size} & \multicolumn{2}{c}{4} & \multicolumn{2}{c}{4} & \multicolumn{2}{c}{4} & \multicolumn{2}{c}{4} & \multicolumn{2}{c}{4} & \multicolumn{2}{c}{4} & \multicolumn{2}{c}{4} & \multicolumn{2}{c}{4} \\
    \midrule
    \multicolumn{2}{c}{Epoch} & \multicolumn{2}{c}{200} & \multicolumn{2}{c}{200} & \multicolumn{2}{c}{200} & \multicolumn{2}{c}{200} & \multicolumn{2}{c}{200} & \multicolumn{2}{c}{100} & \multicolumn{2}{c}{200} & \multicolumn{2}{c}{200} \\
    \midrule
    \multicolumn{2}{c}{Accum.} & \multicolumn{2}{c}{1} & \multicolumn{2}{c}{1} & \multicolumn{2}{c}{1} & \multicolumn{2}{c}{1} & \multicolumn{2}{c}{1} & \multicolumn{2}{c}{1} & \multicolumn{2}{c}{1} & \multicolumn{2}{c}{1} \\
    \midrule
    \multicolumn{2}{c}{Optimizer} & \multicolumn{2}{c}{Adam} & \multicolumn{2}{c}{Adam} & \multicolumn{2}{c}{SGD} & \multicolumn{2}{c}{AdamW} & \multicolumn{2}{c}{AdamW} & \multicolumn{2}{c}{AdamW} & \multicolumn{2}{c}{AdamW} & \multicolumn{2}{c}{Adam} \\
    \midrule
    \multicolumn{2}{c}{Learning Rate} & \multicolumn{2}{c}{1.00E-04} & \multicolumn{2}{c}{1.00E-05} & \multicolumn{2}{c}{1.00E-03} & \multicolumn{2}{c}{1.00E-04} & \multicolumn{2}{c}{1.00E-04} & \multicolumn{2}{c}{1.00E-04} & \multicolumn{2}{c}{1.00E-04} & \multicolumn{2}{c}{1.00E-04} \\
    \midrule
    \multicolumn{2}{c}{Weight Decay} & \multicolumn{2}{c}{1.00E-05} & \multicolumn{2}{c}{1.00E-05} & \multicolumn{2}{c}{1.00E-05} & \multicolumn{2}{c}{0} & \multicolumn{2}{c}{1.00E-03} & \multicolumn{2}{c}{1.00E-05} & \multicolumn{2}{c}{0} & \multicolumn{2}{c}{1.00E-03} \\
    \bottomrule
    \end{tabular}}%
  \label{tab:pdehyper}%
\end{table*}%

\subsection{Detailed results on OpenML-CC18 benchmark} \label{appendix:openml}

The detailed results of \ours and other compared methods are shown in Table~\ref{table:tabular_results_table}, the average accuracy of \ours across 30 datasets is the highest among all methods, with the highest results observed in 15 out of the 30 datasets. Compared to ORCA, the cross-modal fine-tuning approach like us, \ours is better on 22 datasets.

\begin{figure}[htbp]
  \center
  \begin{minipage}{0.8\textwidth}
    \centering
    \includegraphics[width=\textwidth]{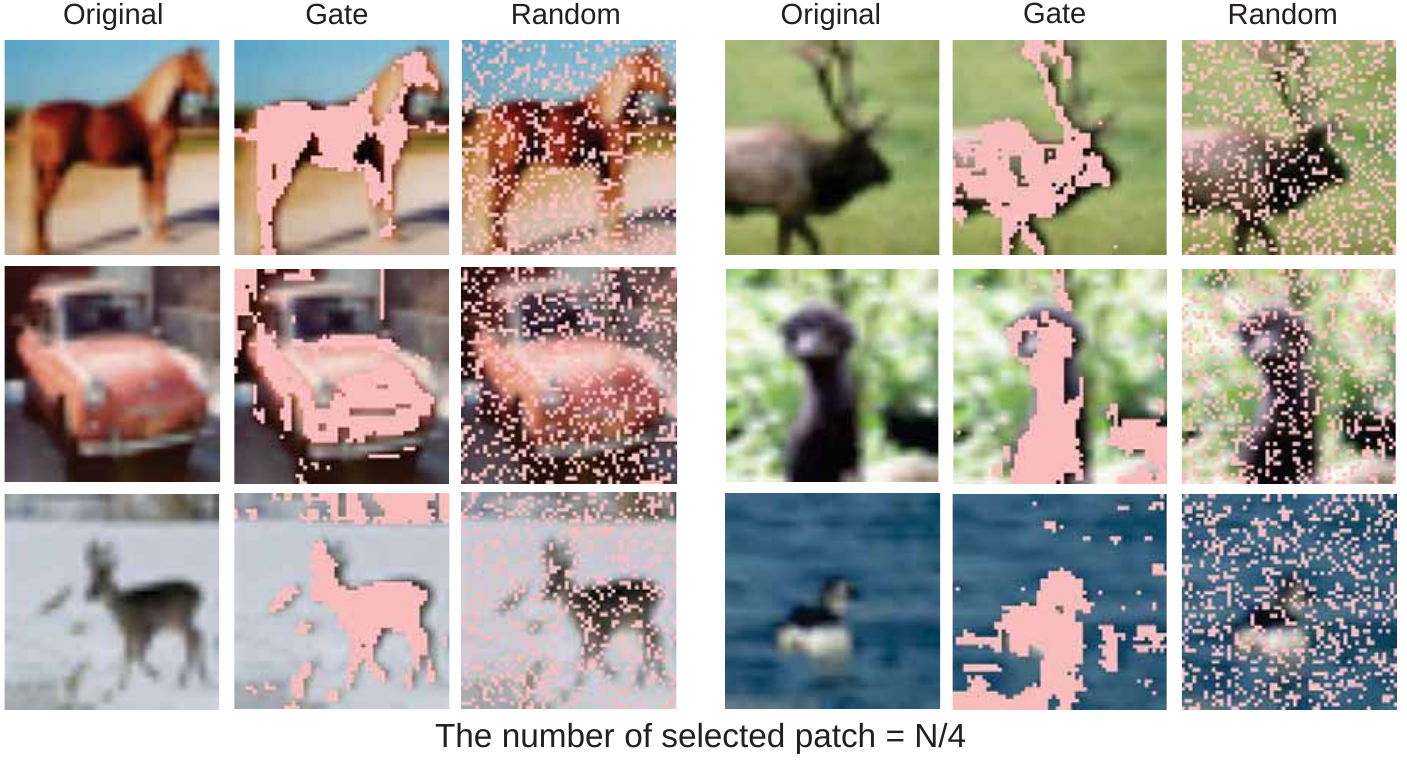}
  \end{minipage}
  \begin{minipage}{0.8\textwidth}
    \centering
    \includegraphics[width=\textwidth]{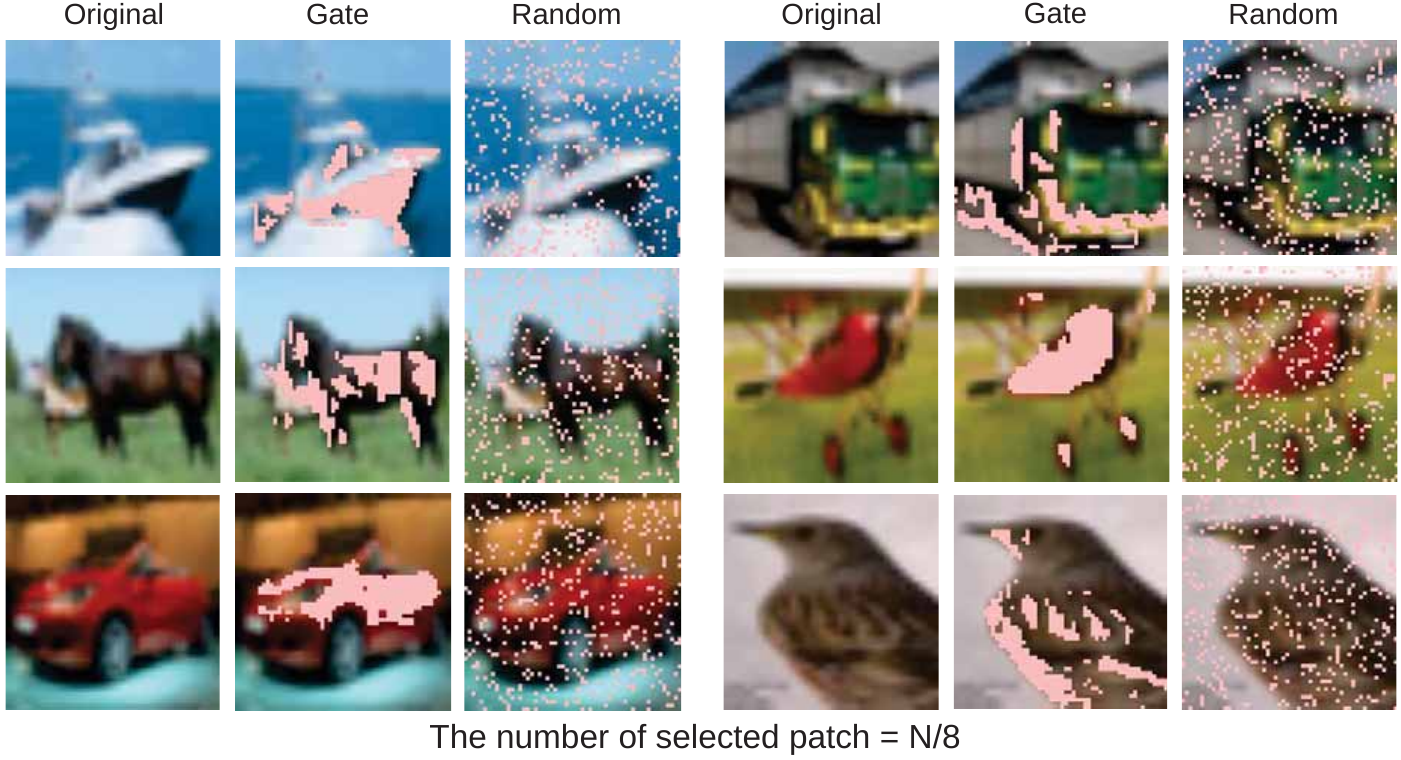}
  \end{minipage}
  \begin{minipage}{0.8\textwidth}
    \centering
    \includegraphics[width=\textwidth]{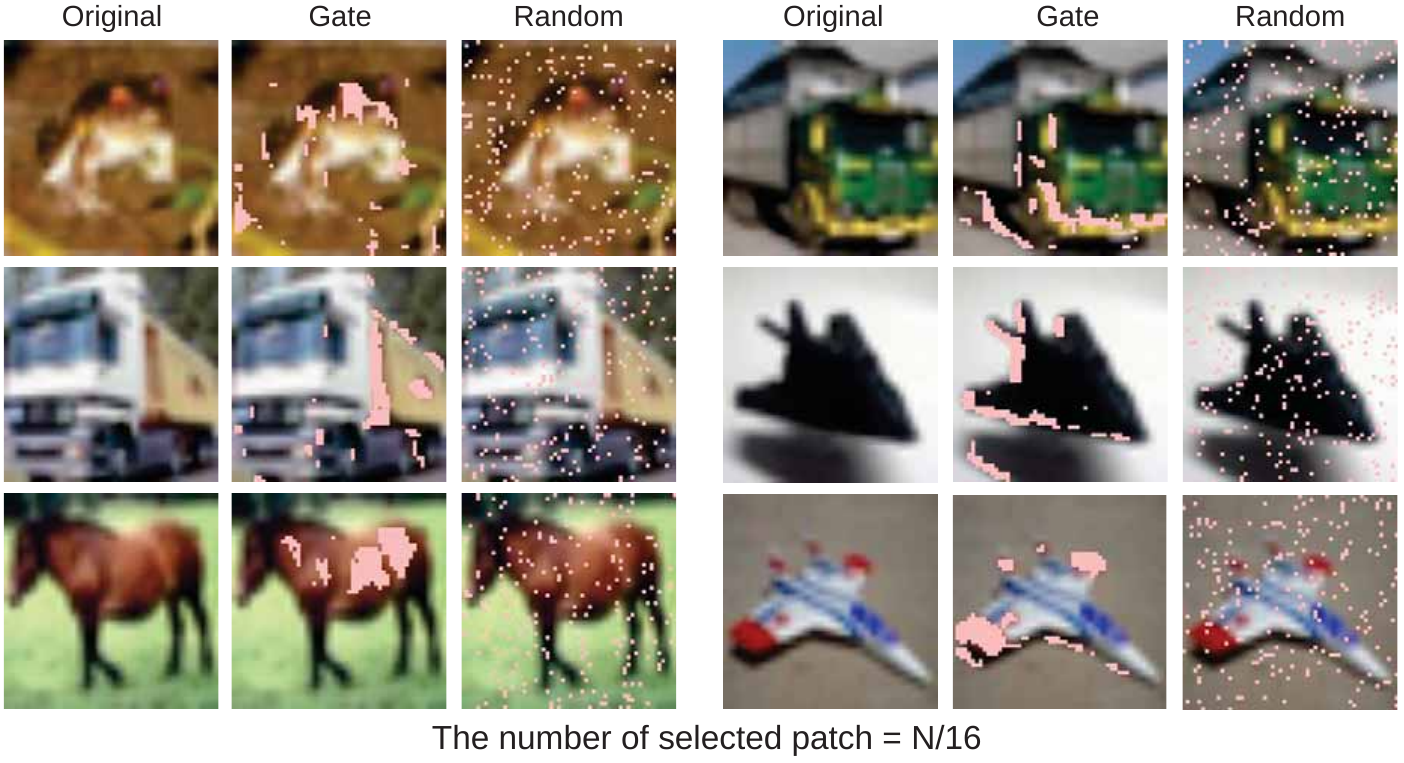}
  \end{minipage}
  \caption{The visualization of the different numbers of patches selected by random strategy and our gate strategy.} \label{fig:gate2}
\end{figure}

\begin{figure}[htbp]
  \center
  \begin{minipage}{0.65\textwidth}
    \centering
    \includegraphics[width=\textwidth]{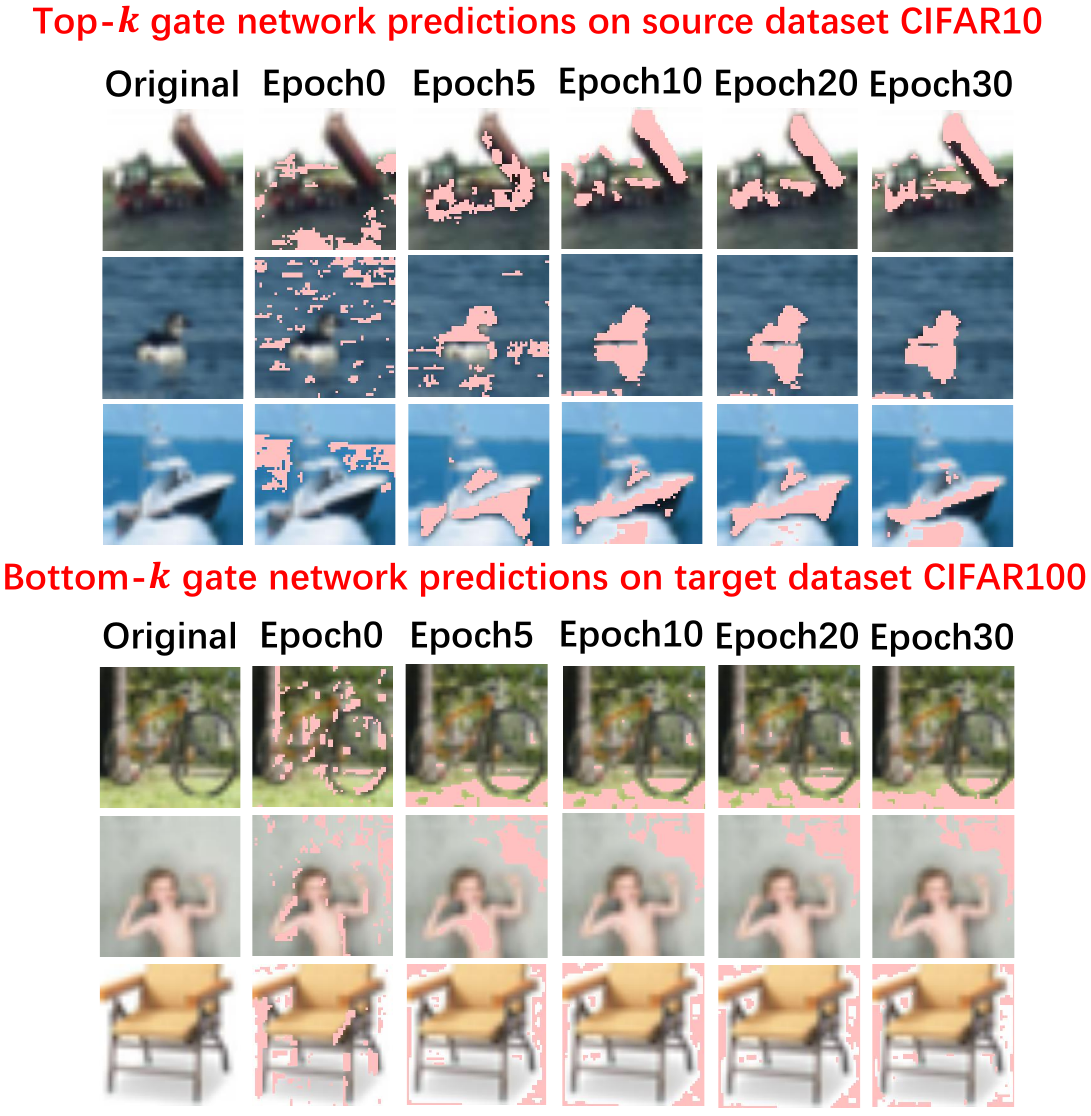}
  \end{minipage}
  \begin{minipage}{0.65\textwidth}
    \centering
    \includegraphics[width=\textwidth]{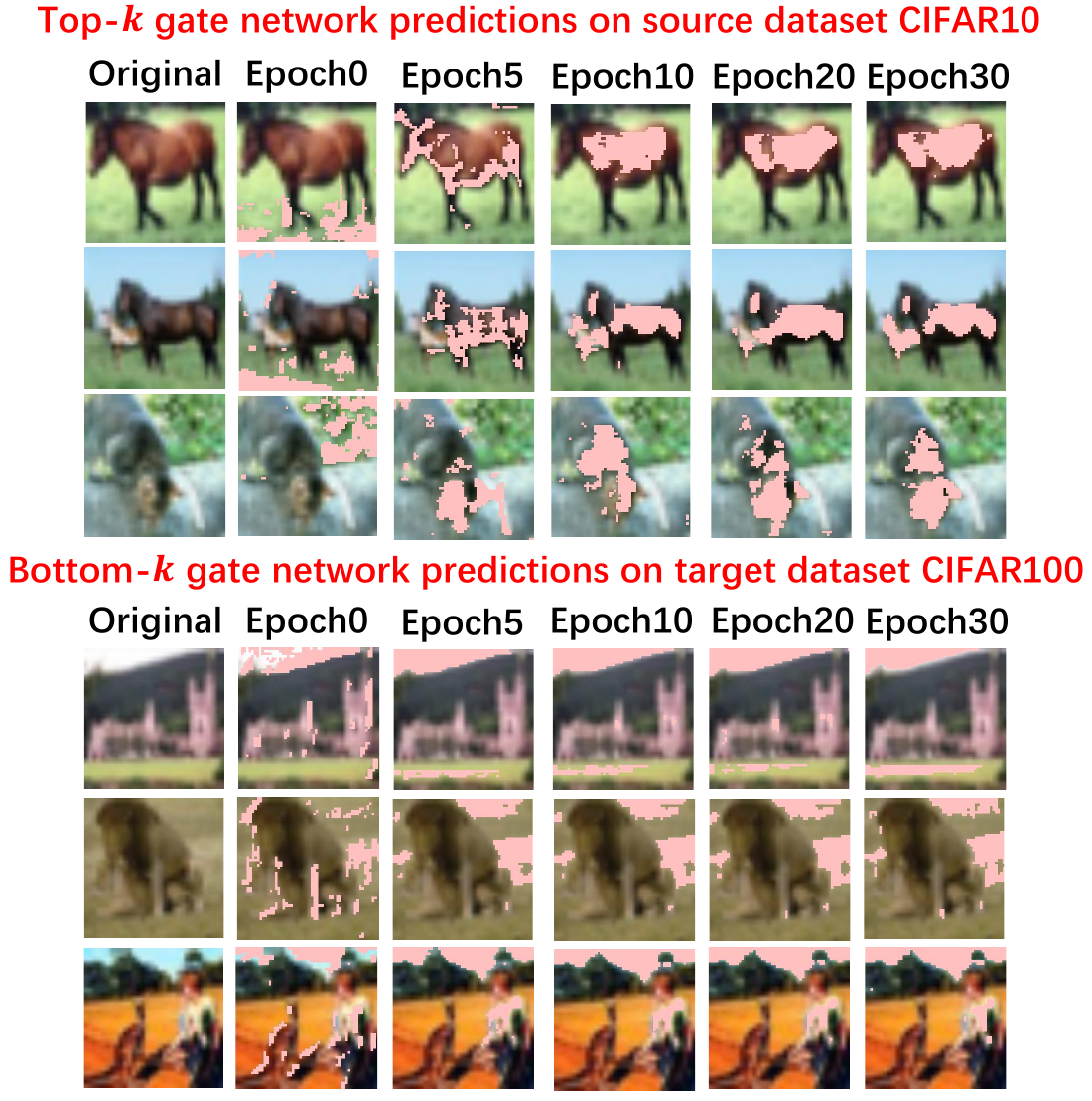}
  \end{minipage}
  \caption{The visualization of the different numbers of patches selected by random strategy and our gate strategy.} \label{fig:gate3}
\end{figure}

\begin{table}[h]
\caption{One-vs-one AUROC ($\uparrow$) on 30 OpenML-CC18 datasets. ``Diff. from XGBoost" is the acrosstask average of per-task difference from XGBoost. On 15/30 datasets, \ours ranks the first among all compared methods.} \vspace{-0.8em}
    \label{table:tabular_results_table}
    \centering
\resizebox{1.0\textwidth}{!}{\begin{tabular}{l|ccc|cc|cc}
\toprule
& LightGBM & CatBoost & XGBoost & AutoGluon &  TabPFN  &ORCA & \cellcolor{bestcolor}\ours\\
\midrule
balance-scale        &          0.9938 &          0.9245 &           0.9939 &                  0.9919   & \textbf{ 0.9973 }  &0.9949 & \cellcolor{bestcolor} 0.9964 \\
mfeat-fourier        &          0.9786 &          0.9816 &           0.9803  &         \textbf{0.9843}   &           0.9811  &0.9729 & \cellcolor{bestcolor}0.9783 \\
breast-w             &           0.991 &          0.9931 &           0.9896  &                  0.9933 &         0.9934  &\textbf{0.9939} &0.9909 \cellcolor{bestcolor}\\
mfeat-karhunen       &          0.9979 &          0.9986 &           0.9983 &      0.9987 &           0.9978   &0.9968 &\textbf{0.9988} \cellcolor{bestcolor}
\\
mfeat-morphologica.. &          0.9601 &          0.9629 &           0.9612 &                  \textbf{0.9698}  &           0.9669 &0.9647 &0.9680  \cellcolor{bestcolor}\\
mfeat-zernike        &          0.9716 &          0.9759 &           0.9735 &         \textbf{0.9908} &           0.9823   &0.9829 &0.9849\cellcolor{bestcolor}
\\
cmc                  &          0.7288 &          0.7256 &           0.7299 &                  0.7331  &           0.7276 &0.7237 &\textbf{0.7770}\cellcolor{bestcolor}
 \\
credit-approval      &          0.9415 &          0.9389 &  0.9422  &                  0.9415&           0.9322 &0.9340 &\textbf{0.9601}\cellcolor{bestcolor}
\\
credit-g             &          0.7684 &          0.7852 &           0.7853  &                  0.7941   &           0.7894  &0.7748 & \textbf{0.8200}\cellcolor{bestcolor}
 \\
diabetes             &          0.8247 &          0.8383 &           0.8378  &                  0.8391   &            0.8410  &0.8239 &\textbf{0.8570 }\cellcolor{bestcolor}\\
tic-tac-toe          &          0.9988 &          0.9992 &       \textbf{1}  &              \textbf{1}   &           0.9759 &0.9973 &0.9952\cellcolor{bestcolor}
 \\
vehicle              &          0.9232 &          0.9302 &           0.9282  &                  0.9416   &  0.9589  &\textbf{0.9591} &0.9582\cellcolor{bestcolor}
\\
eucalyptus           &          0.8931 &          0.8979 &           0.9004 &                  0.9204   &           0.9245 &0.9084 & \textbf{0.9510}\cellcolor{bestcolor}
 \\
analcatdata\_author.. &          0.9999 &          0.9999 &           0.9997   &                  0.9993   &       \textbf{1}  &0.9996 &\textbf{1}\cellcolor{bestcolor}
 \\
analcatdata\_dmft     &          0.5461 &          0.5589 &           0.5743  &                  0.5657  &  \textbf{ 0.579} &0.5627 &0.5509\cellcolor{bestcolor}
 \\
pc4                  &          0.9301 &          0.9413 &           0.9291   &                \textbf{  0.9428}   &           0.9383  &0.9226 &0.9301\cellcolor{bestcolor}\\
pc3                  &          0.8178 &          0.8247 &           0.8288   &                  0.8282   &  0.8373  &0.8411 &\textbf{0.8493}\cellcolor{bestcolor}
\\
kc2                  &          0.8141 &          0.8323 &           0.8227   &                  0.8242 &  0.8346  &\textbf{0.8431} &0.8398\cellcolor{bestcolor}
\\
pc1                  &          0.8321 &            0.86 &           0.8489 &                  0.8578  &           0.8761   &0.8767 & \textbf{0.9266}\cellcolor{bestcolor}\\
banknote-authentic.. &      \textbf{1} &      \textbf{1} &       \textbf{1}   &              \textbf{1}  &       \textbf{1}  &\textbf{1} &\textbf{1} \cellcolor{bestcolor}
 \\
blood-transfusion-.. &          0.7144 &          0.7403 &           0.7312  &                  0.7364  &  0.7549 &\textbf{0.7565} &0.6287\cellcolor{bestcolor}
\\
ilpd                 &          0.6917 &          0.7279 &           0.7171   &                   0.723   &           0.7379  &0.7419 &\textbf{0.7927}\cellcolor{bestcolor}
 \\
qsar-biodeg          &          0.9126 &          0.9217 &           0.9191   &                  0.9276  &         {  0.9336} &\textbf{0.9349} & 0.9167\cellcolor{bestcolor}
 \\
wdbc                 &          0.9904 &          0.9931 &           0.9904   &                  0.9956  &  \textbf{0.9964} &0.9929 &0.9947 \cellcolor{bestcolor}\\
cylinder-bands       &          0.8556 &          0.8757 &           0.8782   &         0.8878  &           0.8336 &0.844 &\textbf{0.9243}\cellcolor{bestcolor}
 \\
dresses-sales        &          0.5593 &          0.5696 &  0.5823 &                  0.5507  &           0.5376   &\textbf{0.6025} & 0.5747\cellcolor{bestcolor}
 \\
MiceProtein          &          0.9997 &          0.9999 &           0.9998  &              \textbf{1}  &           0.9999 &0.9969 & 0.9997\cellcolor{bestcolor}
\\
car                  &          0.9925 &          0.9955 &           0.9948 &   0.998   &            0.995 &0.9983 & \textbf{1}\cellcolor{bestcolor}
 \\
steel-plates-fault.. &          0.9626 &          0.9655 &           0.9656  &                  0.9666   &           0.9655  &0.9543 & \textbf{0.9696}\cellcolor{bestcolor}
\\
climate-model-simu.. &          0.9286 &          0.9344 &           0.9255  &                  0.9391  &           0.9415  &0.9416 & \textbf{0.9551} \cellcolor{bestcolor} \\
\hline
\# Wins   &   1	&1	&2	&7	&5&	7 & \textbf{15}  \cellcolor{bestcolor}\\
\hline
Avg. AUROC 
                                         &  0.8840 &   0.8898 &   0.8909 &              0.8947  &    0.8943&   0.8946 & \textbf{0.9030}\cellcolor{bestcolor}\\
\hline
Avg. Diff. from XGBoost &  -0.0069&   -0.0011 &   0 &  +0.0038  &  +0.0034 &   +0.0036 & \textbf{+0.0121}\cellcolor{bestcolor}\\
\bottomrule
\end{tabular}}
\vspace{-0.8em}
\end{table}

\subsection{Visualization of patch selection} \label{appendix:pase}

We visualize the patches selected by random strategy and our gate strategy in Fig.~\ref{fig:gate2}. The right column in Fig.~\ref{fig:gate2} highlights patches selected randomly, visualized in red, we notice certain drawbacks. The random selection may result in an overabundance of background patches and a lack of continuity in the critical patches, potentially impeding the model's ability to capture essential data information and disrupting its training process. To address this issue, we introduce a gate network for patch scoring. As depicted in Fig.~\ref{fig:gate2}, the middle column illustrates patches selected through patch scoring. Notably, patches chosen through this method demonstrate a heightened focus on the primary components of the data, facilitating more comprehensive training of the model on the intermediate modality.

Furthermore, we illustrate the evolution of the gate network's predictions as it learns to integrate information from both the source modality (CIFAR10) and target modality (CIFAR100) in Fig.~\ref{fig:gate3}. As depicted in the visualization, we observe that initially, the gate network tends to produce random predictions. However, as training progresses, typically around the 10th epoch, we notice a significant transition as the gate network approaches convergence. At this stage, the gate network demonstrates a clear ability to select critical patches of the source modality data and the unimportant patches of the target modality data. After patch replacement, we can effectively maximize the retention of information from both modalities.

\subsection{Additional evaluation on NAS-Bench-360} \label{appendix:nas}

\subsubsection{Comparison with different cross-modal fine-tuning method}

In Table~\ref{tab:cmft}, we compare our method, \ours, with various cross-modal fine-tuning approaches, including Train\_from\_scratch, FPT (layernorm), NFT (naive full fine-tuning), and ORCA. The results indicate that our method outperforms all other cross-modal fine-tuning methods across all tasks.

\begin{table*}[htbp]
  \centering
  \caption{Comparison with different cross-modal fine-tuning method including: Train\_from\_scratch: training SwinTransformer/RoBERTa from scratch; FPT: fine-tuning only the layernorm; NFT: fine-tuning all parameters, ORCA and \ours.}
  \resizebox{0.95\textwidth}{!}{
    \begin{tabular}{cccccccccccccccccccccc}
    \toprule
    & CIFAR-100 & Spherical & Darcy Flow & PSICOV & Cosmic & NinaPro & FSD50K & ECG & Satellite & DeepSEA \\
    \midrule
    Train\_from\_scratch & 50.87 & 76.67 & 8.00E-02 & 5.09 & 0.5 & 9.96 & 0.75 & 0.42 & 12.38 & 0.39 \\
    \midrule
    FPT & 10.11 & 76.38 & 2.10E-02 & 4.66 & 0.233 & 15.69 & 0.67 & 0.5 & 20.83 & 0.37 \\
    NFT & 7.67 & 55.26 & 7.34E-03 & 1.92 & 0.17 & 8.35 & 0.63 & 0.44 & 13.86 & 0.51 \\
    ORCA & 6.53 & 29.85 & 7.28E-03 & 1.91 & 0.152 & 7.54 & 0.56 & \textbf{0.28} & 11.59 & 0.29 \\
    \cellcolor{bestcolor}\ours & \cellcolor{bestcolor}\textbf{6.25} & \cellcolor{bestcolor}\textbf{25.55} & \cellcolor{bestcolor}\textbf{7.00E-03} & \cellcolor{bestcolor}\textbf{0.99} & \cellcolor{bestcolor}\textbf{0.121} & \cellcolor{bestcolor}\textbf{6.53} & \cellcolor{bestcolor}\textbf{0.55} & \cellcolor{bestcolor}\textbf{0.28} & \cellcolor{bestcolor}\textbf{11.18} & \cellcolor{bestcolor}\textbf{0.28} \\
    \bottomrule
    \end{tabular}}%
  \label{tab:cmft}%
\end{table*}%

\subsubsection{In-modality transfer}
In the preceding sections, we have thoroughly demonstrated the superiority of our method in the context of cross-modal fine-tuning. Next, we will validate the effectiveness of our approach in in-modality transfer in Table~\ref{tab:domainnet}. Taking the miniDomainNet~\cite{ZhouYQX21_minidomainnet} (a reduced version
of DomainNet~\cite{DomainNet}) dataset with 126 classes as an example, we will assess the effectiveness of our method for in-modality transfer on four significantly different domains (Clipart, Painting, Real and Sketch)

\begin{table*}[htbp]
  \centering
  \caption{Prediction errors ($\downarrow$) on four domains on miniDomainNet.}
    \begin{tabular}{ccccc}
    \toprule
    & Clipart & Painting & Real & Sketch \\
    \midrule
    ORCA & 10.16 & 12.86 & \textbf{5.08} & 14.29 \\
    \cellcolor{bestcolor}\ours & \cellcolor{bestcolor}\textbf{9.21} & \cellcolor{bestcolor}\textbf{11.75} & \cellcolor{bestcolor}\textbf{5.08} & \cellcolor{bestcolor}\textbf{14.13} \\
    \bottomrule
    \end{tabular}%
  \label{tab:domainnet}%
\end{table*}%

\subsubsection{More data-limited scenarios}
What performance differences arise when our method is applied to target modalities with scarcer data? To investigate this question, we further reduce the data in the target modality and compare the results with ORCA on three datasets: 2D classification Ninapro and 1D classification DeepSEA. We compare the results in Table~\ref{tab:limit} using 10\%, 30\%, 50\%, 70\%, and 90\% of the training data. We found that even with less training data, our method still achieves better performance compared to ORCA.

\begin{table*}[htbp]
  \centering
  \caption{The results of more limited data in target modality.}
    \begin{tabular}{ccccccc}
    \toprule
    Ninapro & 10\%  & 30\%  & 50\%  & 70\%  & 90\% \\
    \midrule
    ORCA & 27.31 & 16.39 & 12.75 & 10.02 & 9.41 \\
    \cellcolor{bestcolor}\ours & \cellcolor{bestcolor}\textbf{18.82} & \cellcolor{bestcolor}\textbf{15.17} & \cellcolor{bestcolor}\textbf{12.14} & \cellcolor{bestcolor}\textbf{8.04} & \cellcolor{bestcolor}\textbf{7.59} \\
    \midrule
    DeepSEA & 10\%  & 30\%  & 50\%  & 70\%  & 90\% \\
    \midrule
    ORCA & 0.369 & 0.355 & 0.349 & 0.345 & 0.339 \\
    \cellcolor{bestcolor}\ours & \cellcolor{bestcolor}\textbf{0.363} & \cellcolor{bestcolor}\textbf{0.353} & \cellcolor{bestcolor}\textbf{0.348} & \cellcolor{bestcolor}\textbf{0.327} & \cellcolor{bestcolor}\textbf{0.286} \\
    \bottomrule
    \end{tabular}%
  \label{tab:limit}%
\end{table*}%

\subsubsection{Comparison with Data-augmentation}

Data augmentation is an essential approach to enrich dataset diversity and prevent model overfitting. However, due to the diverse nature of target modalities, it's challenging to apply uniform data augmentation techniques across all modalities. For instance, common image augmentation techniques like random crop or grayscale may not be intuitively applicable to other modalities such as PDEs or cosmic rays and might even mislead the model. 

Therefore, in Table~\ref{tab:aug}, we further explore techniques like mixing up target data or utilizing masking operations for data augmentation. From the results, we observe that this augmentation approach is effective for datasets closer to the source modality, such as CIFAR100 and Spherical, yielding promising results. However, for tasks with significant differences from the source modality, like Darcy Flow or Ninapro, traditional techniques are ineffective and might even have negative impacts. We also find that if we only classify the mix embedding into the target label without incorporating the source loss, it essentially resembles the mask operation and cannot achieve modality-agnosticism. Hence, we ultimately adopt this intermediate modality generation approach for cross-modal transfer. However, designing a modality-agnostic data augmentation approach remains a future research direction.

\begin{table*}[htbp]
  \centering
  \caption{Comparison with in-modality data augmentation strategy.}
    \begin{tabular}{lcccc}
    \toprule
     & CIFAR100 & Spherical & Darcy Flow & Ninapro \\
    \midrule
    target\_mixup & 6.30 & \textbf{24.12} & 8.80E-03 & 9.71 \\
    target\_mask & 6.48 & 27.37 & 7.40E-03 & 7.89 \\
    \ours w/o src\_loss & 6.54 & 26.65 & 7.40E-03 & 8.04 \\
    \cellcolor{bestcolor}\ours & \cellcolor{bestcolor}\textbf{6.25} & \cellcolor{bestcolor}25.55 & \cellcolor{bestcolor}\textbf{7.00E-03} & \cellcolor{bestcolor}\textbf{6.53} \\
    \bottomrule
    \end{tabular}%
  \label{tab:aug}%
\end{table*}%



\end{document}